\documentclass{article}
\usepackage{times}
\usepackage{xcolor}
\usepackage{amsmath, amssymb, xspace, mathalfa}
\usepackage{soul}
\usepackage{amsmath}
\usepackage{graphicx}
\usepackage{epstopdf}
\usepackage[utf8]{inputenc}
\usepackage[small]{caption}
\usepackage{subfig}
\usepackage{algorithm}
\usepackage[noend]{algpseudocode}
\usepackage{tikz}
\usepackage{amssymb}

\def\G{\ensuremath{{\mathcal G}}\xspace}
\def\Pr{\ensuremath{\mathbb{P}}}
\def\programspace{\ensuremath{\mathcal{P}}}
\def\programspacebar{\ensuremath{\mathcal{\bar{P}}}}
\def\canvasspace{\ensuremath{{\mathcal T}}}

\def\testp{\ensuremath{p}}
\def\testc{\ensuremath{c}}

\def\testt{\ensuremath{t}}
\def\testsetsize{\ensuremath{n}}

\def\alg{\mathcal{A}}
\def\interp{\ensuremath{I}}
\def\Ind{\ensuremath{\mathbb{I}}}

\title{Program Synthesis from Visual Specification}

\author{
Evan Hernandez$$, 
Ara Vartanian$$, 
Xiaojin Zhu$$ 
\\ 
$$ University of Wisconsin-Madison \\
hernandez@cs.wisc.edu,
aravart@cs.wisc.edu,
jerryzhu@cs.wisc.edu
}

\begin{document}

\maketitle

\begin{abstract}
  Program synthesis is the process of automatically translating a specification
  into computer code. Traditional synthesis settings require a formal, precise
  specification. Motivated by computer education applications where a student
  learns to code simple turtle-style drawing programs, we study a novel
  synthesis setting where only a noisy user-intention drawing is specified. This
  allows students to sketch their intended output, optionally together with
  their own incomplete program, to automatically produce a completed program. We
  formulate this synthesis problem as search in the space of programs, with the
  score of a state being the Hausdorff distance between the program output and
  the user drawing. We compare several search algorithms on a corpus consisting
  of real user drawings and the corresponding programs, and demonstrate that our
  algorithms can synthesize programs optimally satisfying the specification.
\end{abstract}

\section{Introduction}

The problem of program synthesis is an important one in AI. Synthesis has many
settings, from the fully automated settings, where code is synthesized at the
level of machine code, to the interactive, where synthesis assists a
professional developer in an IDE. We focus on a novel synthesis setting where
synthesis may be used to facilitate computer science education.

\begin{figure}[htb]
\begin{center}
\begin{tabular}{c c c}
   \includegraphics[width=0.1275\textwidth]{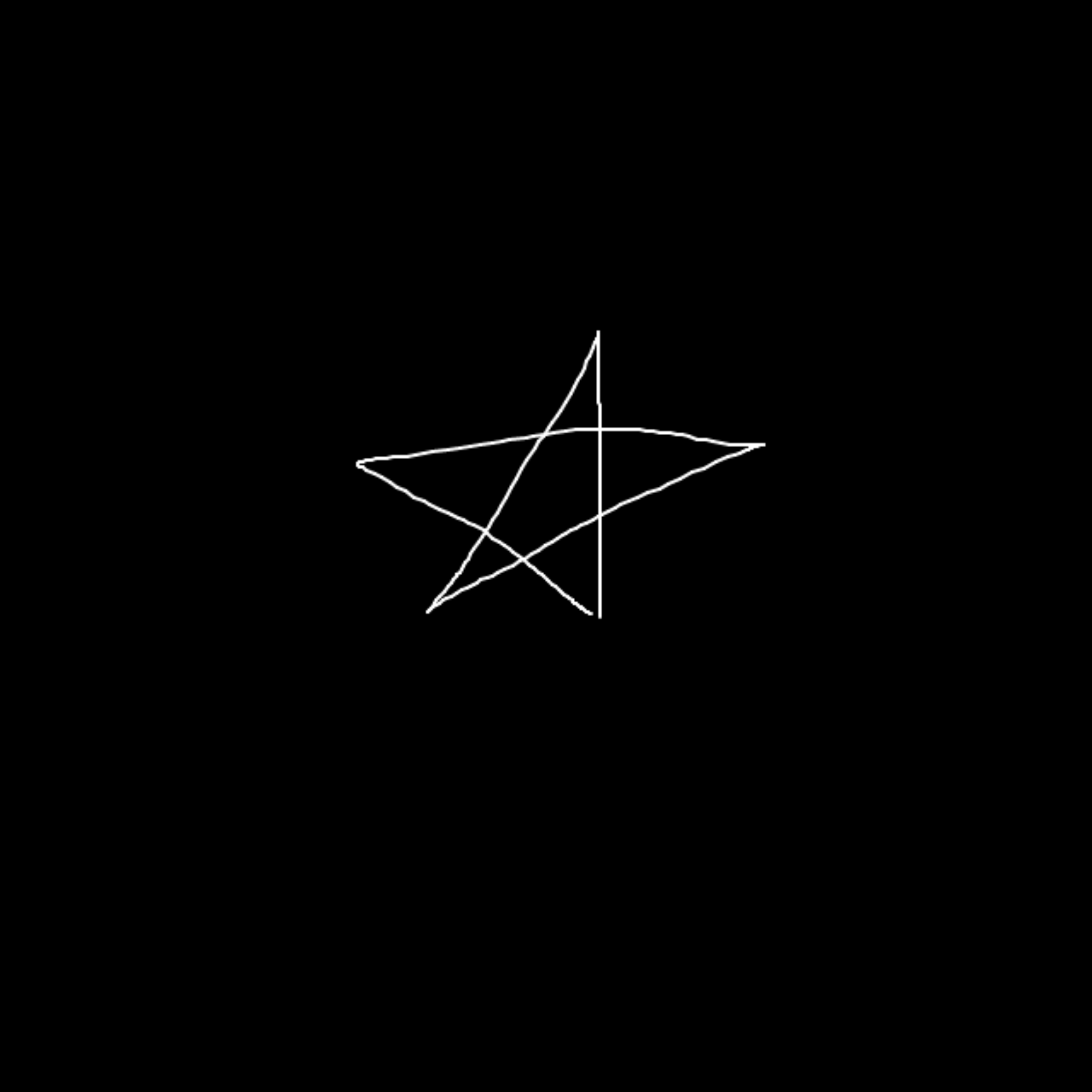} & \includegraphics[width=0.1\textheight]{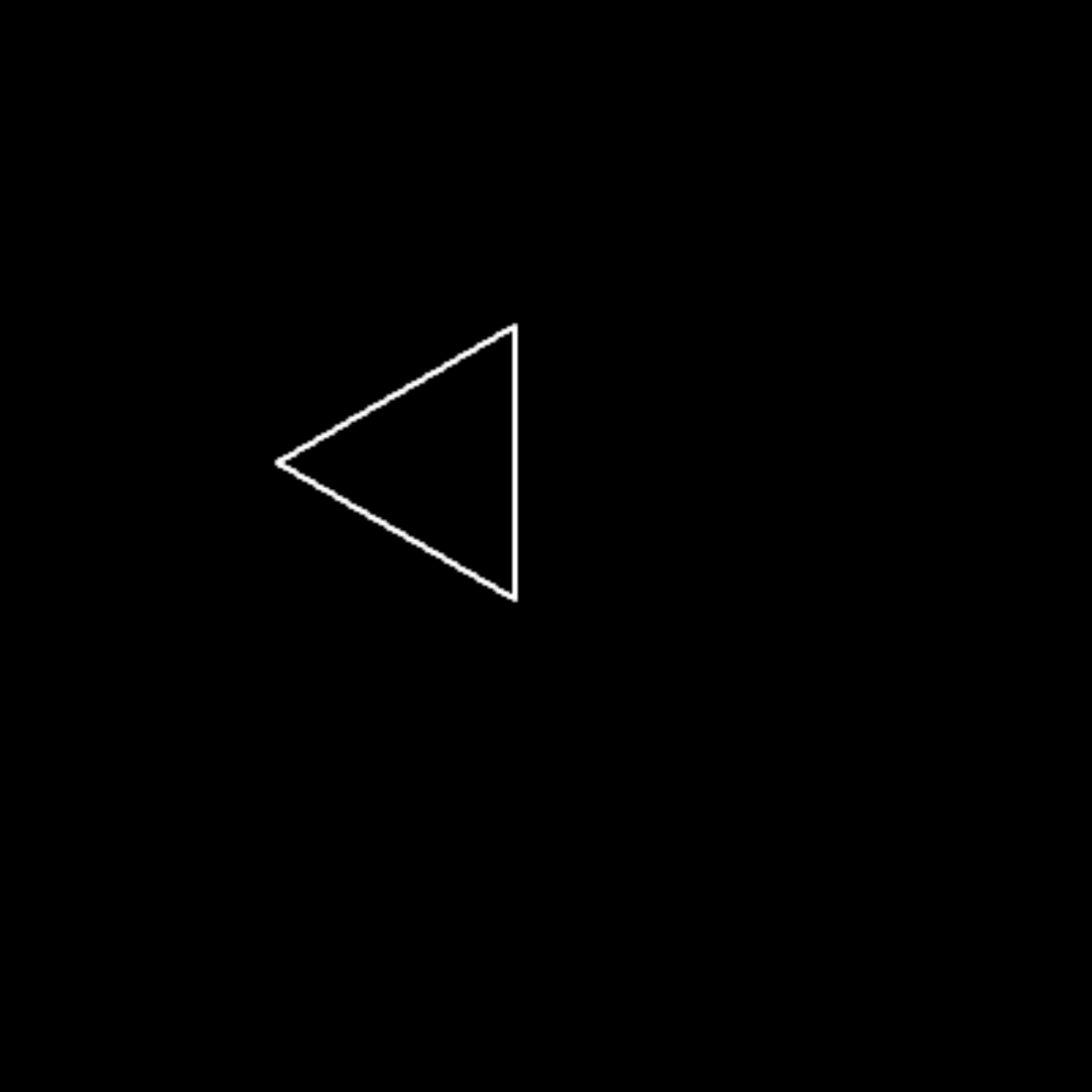} & \includegraphics[width=0.1\textheight]{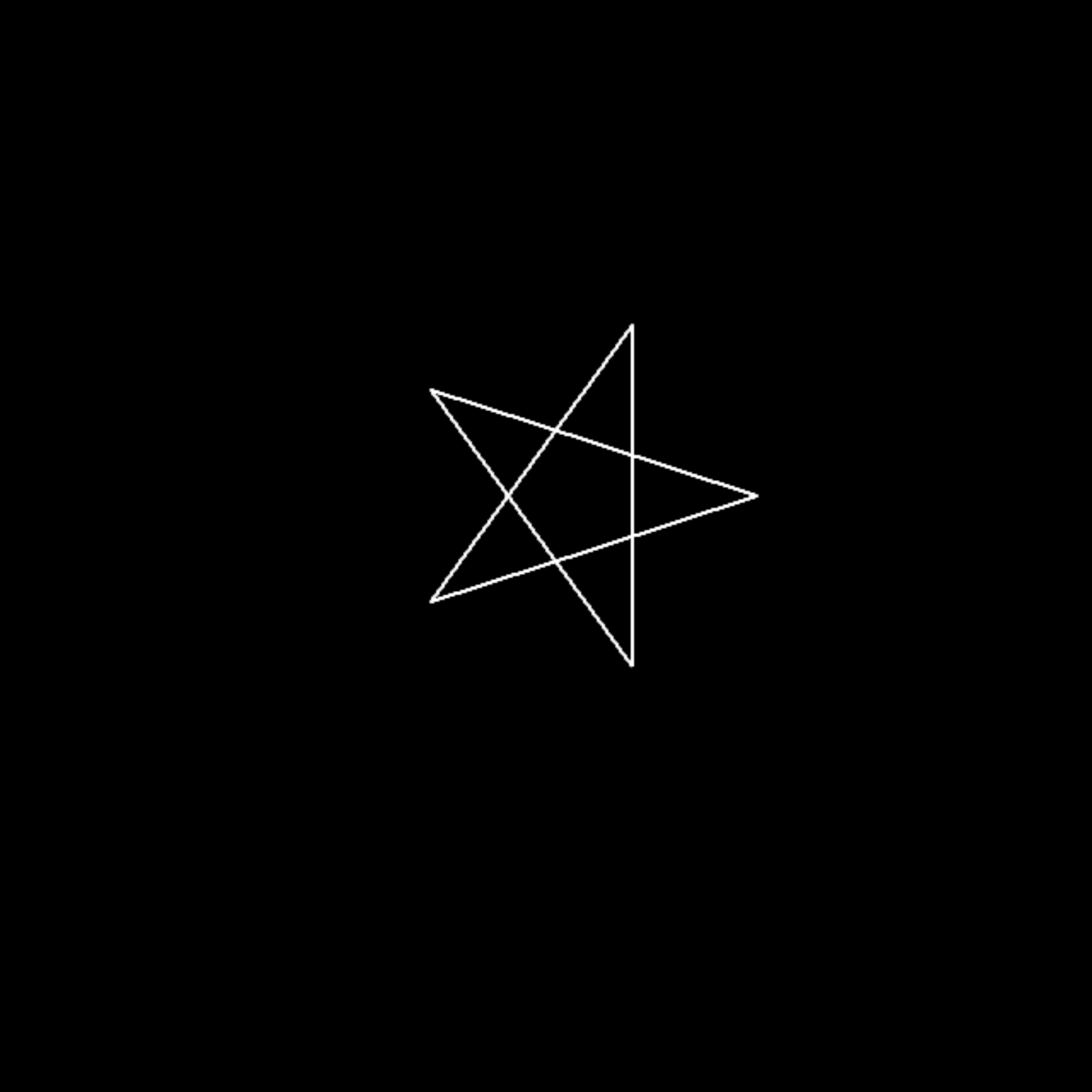} \\
 & \includegraphics[width=0.14\textheight]{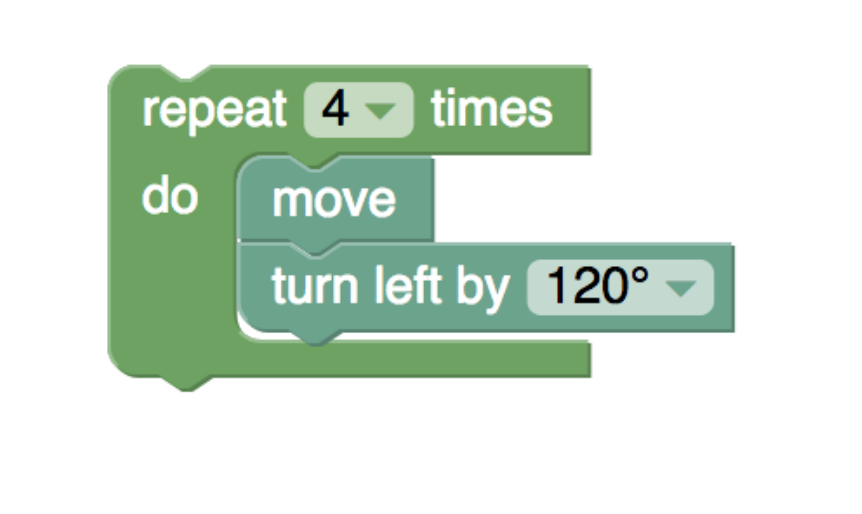} & \includegraphics[width=0.14\textheight]{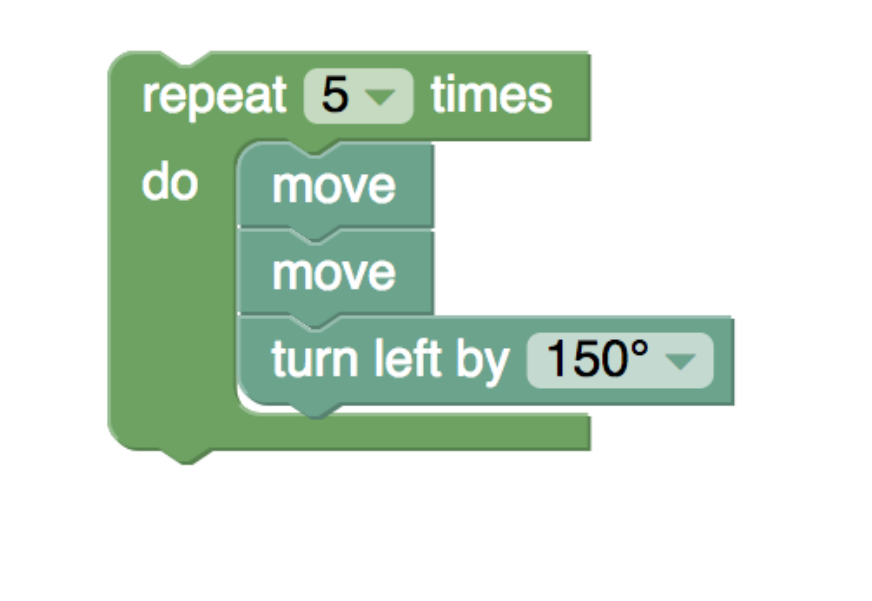} \\
(a) An intended trajectory. & (b) A partial  program. & (c) The synthesized solution. \\
\end{tabular}
\end{center}
\caption{}
\label{fig:star}
\end{figure}

Consider a student in an educational programming task, drawing an image with a
turtle-style program. They may have an intention expressible as a trajectory
they would like to draw, as shown in \textbf{Figure}~\ref{fig:star} (a). This trajectory
can be considered a \emph{complete} but \emph{noisy} specification of the
intended program. And they may have some program built up in their workspace,
but this program may be ``incomplete'' or ``buggy''. For example,
\textbf{Figure}~\ref{fig:star} (b) shows a program the user might have composed with the
intention of drawing \textbf{Figure}~\ref{fig:star} (a), along with the trajectory it
actually creates. The user might be uncertain how to proceed in order to correct
this. It is in this setting where we seek to formulate a synthesis task that can
provide the solution shown in \textbf{Figure}~\ref{fig:star} (c) where the arguments to
both the \texttt{repeat} and \texttt{turn} blocks are corrected and a
\texttt{move} block is added to yield the program nearest to the intended
trajectory.

In this paper, we describe how to synthesize code from such user-provided
visual specification. We ultimately formulate the synthesis problem as an optimization problem
suitable for combinatorial search.

\section{Setting}

\subsection{Programming Language}

We will consider the space $\programspacebar$ of turtle programs
generated by the following grammar:
\begin{align*}
\textit{Program} \rightarrow \ &\textit{Statement}* \\
\textit{Statement} \rightarrow \ &\textsc{Move}  \\ 
				   &| \ \textsc{Turn}\,\textit{Angle} \\
				   &| \ \textsc{Repeat } \textit{Int Program}
\end{align*}

Here $\textit{Angle}$ takes on values at increments of 30 degrees
and $\textit{Int}$ takes on values from 2 to 5. Throughout the paper we will
speak of elements of this space equivalently as either programs or blocks. As can
be seen in \textbf{Figure}~\ref{fig:star} and elsewhere, blocks may be connected by being
nested horizontally within \texttt{repeat} statements or vertically. If a block
is connected vertically beneath another, we refer to it as a \emph{child} block
of its \emph{ancestor}. If a block has no ancestors, we refer to it as a
\emph{root} block.

\begin{figure}[htb]
\begin{center}
\begin{tabular}{c c}
  (a) Two programs. & (b) One program. \\
\includegraphics[width=0.2\textwidth,trim={0 5mm  0 5mm},clip]{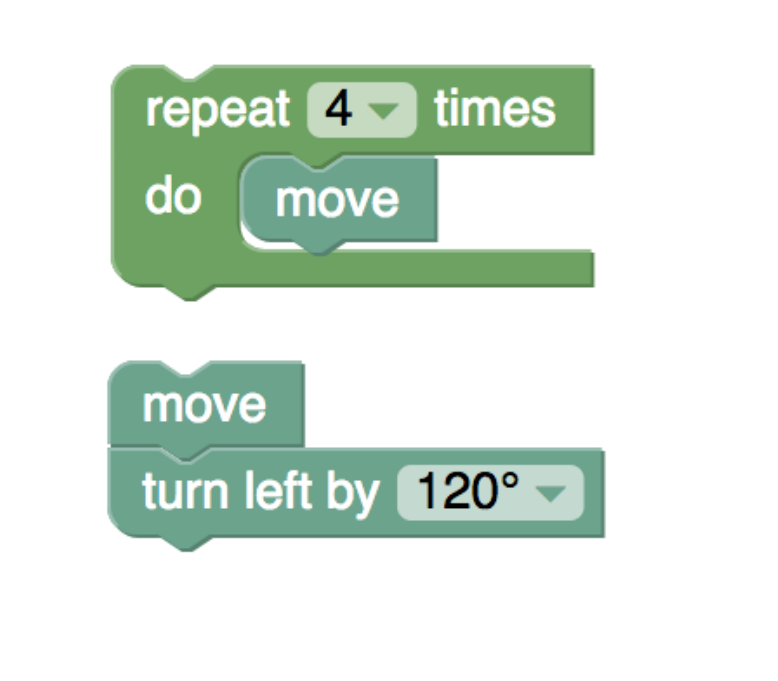} & \includegraphics[width=0.17\textwidth]{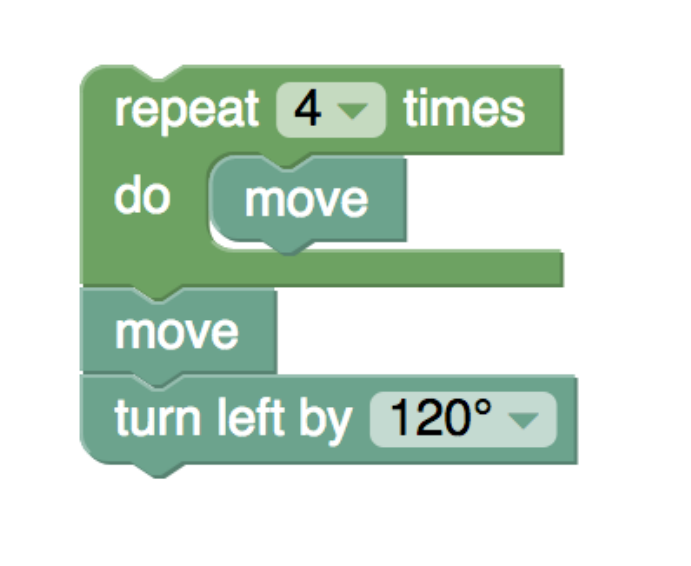}
\end{tabular}
\end{center}
\caption{Examples of workspaces.}
\label{fig:workspace}
\end{figure}

We implement this turtle language using the
Blockly\footnote{https://developers.google.com/blockly} visual block programming
language and its editor. The semantics of this language are as follows:
\begin{itemize}
\item{A \textsc{move} statement translates the turtle in its current direction by some fixed magnitude.}
\item{A \textsc{turn} statement rotates the turtle by the specified number of degrees.}
\item{A \textsc{repeat} statement executes a subprogram some number of times.}
\end{itemize}

A user may position several such elements of $\programspacebar$ on their
\textbf{workspace} as shown in \textbf{Figure}~\ref{fig:workspace} (a). Taking a more
abstract perspective, we can define a workspace as a list of elements in
$\programspacebar$ and we denote the space of workspaces by $\programspace$. We
call a set of points on the two-dimensional plane $t \in \canvasspace$ a
\textbf{trajectory}. And we write $\interp(p) = t$ for the interpretation
function which maps a workspace $p$ to its trajectory $t$. This interpretation
function can be thought of as executing each block in the workspace on the
canvas in the order that it appears in workspace list.

\subsection{Editing Environment} \label{sec:editing}

We can describe the user interface of an editor for our programming language
through \textbf{editing commands}. These commands represent discrete mouse
manipulations performable through Blockly. Note that for these commands to make
sense we must label each block on the workspace with an identifying number.

The families of commands are:

\begin{figure}[H]
\begin{enumerate}
  \item Get a \emph{\{Type\}} block.
  \item Remove block \emph{\{BlockId\}}.
  \item Connect block \emph{\{BlockId\}} under block \emph{\{BlockId\}}.
  \item Connect block \emph{\{BlockId\}} inside block \emph{\{BlockId\}}.
  \item Disconnect block \emph{\{BlockId\}}. 
  \item Change \emph{\{Val\}} in block \emph{\{BlockId\}} to \emph{\{Val\}}.
\end{enumerate}
\caption{Family of available editing commands.}
\label{fig:commands}
\end{figure}

(1) adds a new block to the workspace, as a root block not connected
to any other block on the workspace. The type parameter can be one of
Move, Turn, or Repeat.

(2) removes the block and all its child blocks. This command matches
the Blockly semantics of dragging a block to the trash bin.

(3) and (4) move a block and all its children to a new location. (3) moves a source block
under a target block and connects them. If the target block has children, they are appended
under the source block's children. (4) is distinguished from (3) in that the target must be a repeat
block, and it places the source block at the top of the repeat body.

(5) disconnects a block from its parent, making it into a root block
on the workspace.

(6) modifies the parameter values of blocks, such as the angle in the turn block
or the integer in the repeat block.

The user writes a program by applying a sequence of editing commands beginning from an empty
workspace. That is, a command, when specialized by a choice of feasible values
for its parameters, can be thought of as mapping a workspace $p$ to a successor
workspace $p'$. For example, beginning at an empty workspace, the following sequence
of commands produces the program in \textbf{Figure}~\ref{fig:star} (b):
\begin{enumerate}
  \item Get a repeat block.
  \item Get a move block.
  \item Connect block 2 inside block 1.
  \item Get a turn block.
  \item Connect block 3 under block 2.
  \item Change 30 in block 3 to 120.
\end{enumerate}
This family of commands represents an abstraction of the editor's capabilities,
as the user would typically be manipulating the editing environment with
keyboard and mouse.

\section{Problem Formulation}

Having described the programming language and editing environment, we are now in
a position to formulate our synthesis problem as search. The user in our a
programming environment intends to produce a trajectory $t^* \in \canvasspace$
by means of a turtle program. Consider this as a search on a graph $\G$ whose
vertices are the set of workspaces $\programspace$. There is an edge from $p$ to
$p'$ in $\G$ if there is an editing command which produces workspace $p'$ when
applied to workspace $p$. All edges have unit cost. We let $\textrm{cost}(p,p')$
designate the weight of the shortest path from $p$ to $p'$ in $\G$. We will
designate the initial state of their workspace by $p_0$.

We measure similarity of trajectories with Hausdorff distance.
The Hausdorff distance is a commonly used metric for tasks in object
matching and image analysis \cite{jesorsky2001robust} and serves as a natural
metric for the quality of the fit of a candidate program to a trajectory. We
denote the Hausdorff distance between sets of points $X$ and $Y$ by
  $$d_H(X,Y) := \max(\max_{x \in X} \min_{y \in Y} d(x,y), \max_{y \in Y} \min_{x \in X}
  d(x,y))$$ where $d(\cdot,\cdot)$ is the ordinary Euclidean distance.

Synthesizing a good solution $\hat{p}$ from $p_0$ for trajectory $t^*$
involves a tradeoff between two types of distance or error. On the one hand, a
candidate solution $p$ has some distance from the target trajectory
$d_H(I(p),t^*)$, reflecting the quality of its fit. On the other hand, it may
depart significantly from $p_0$, reflected in a large value of
$\textrm{cost}(p_0,p)$, the minimum number of editing steps required to create
$p$ if beginning from $p_0$.

We trade these off in a constraint formulation. Given intended trajectory
$t^*$ and current program $p_0$, we define our synthesis problem as:

\begin{equation*}
\begin{aligned}
& \underset{p \in \programspace}{\text{argmin}}
& & d_H(\interp(p), t^*) \\
& \,\,\,\,\,\,\text{st}
& & \textrm{cost}(p,p_0) \le C.
\end{aligned}
\end{equation*}

In all experiments below, we choose $C = 6$. This choice was made to ensure
practical run times.

\section{Algorithms}

\subsection{IDPS}

\begin{figure}
\begin{center}
\begin{tabular}{c c c}
\includegraphics[width=0.11\textheight,height=0.11\textheight,trim={0 2cm 3cm 0},clip]{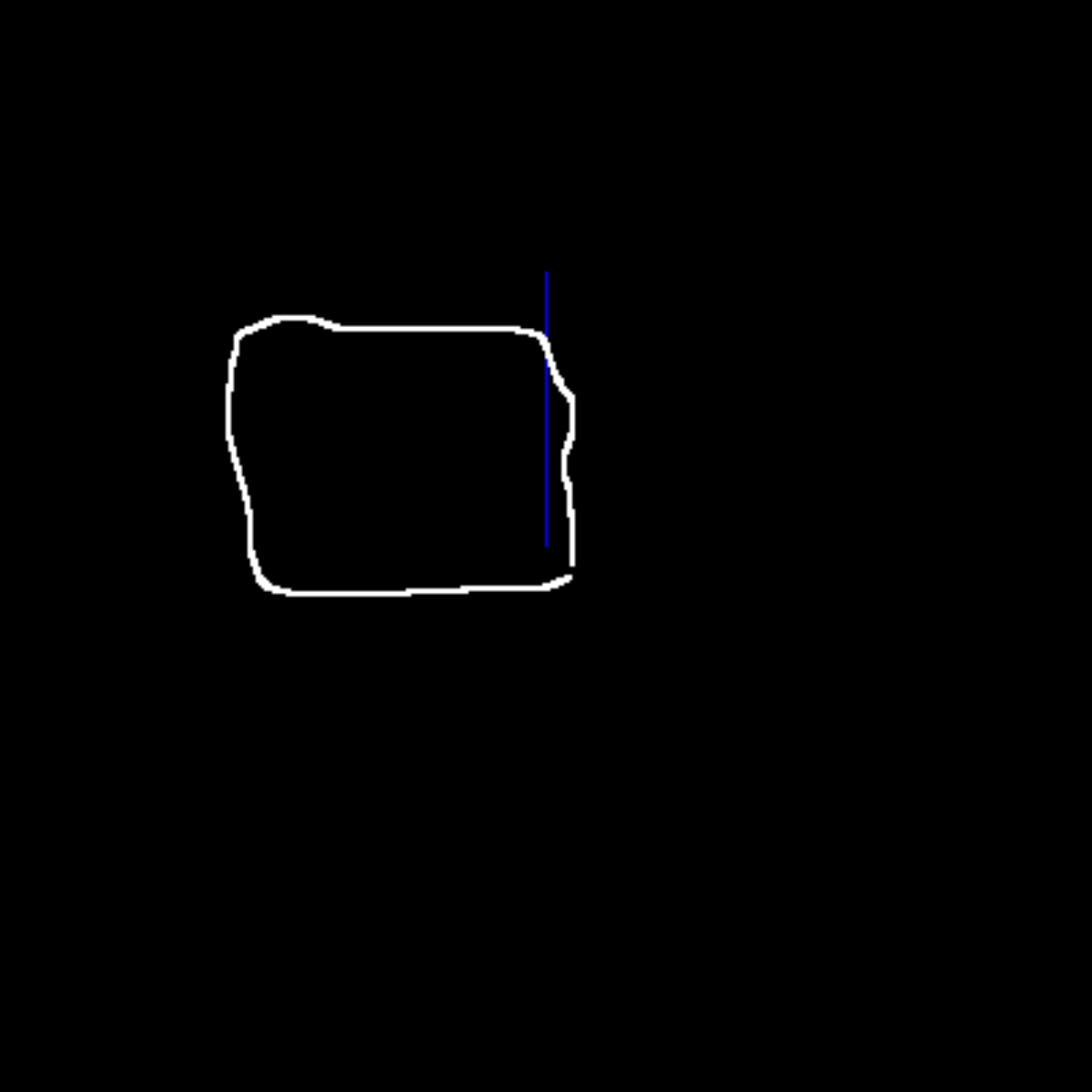} &
\includegraphics[width=0.11\textheight,height=0.11\textheight,trim={0 2cm 3cm 0},clip]{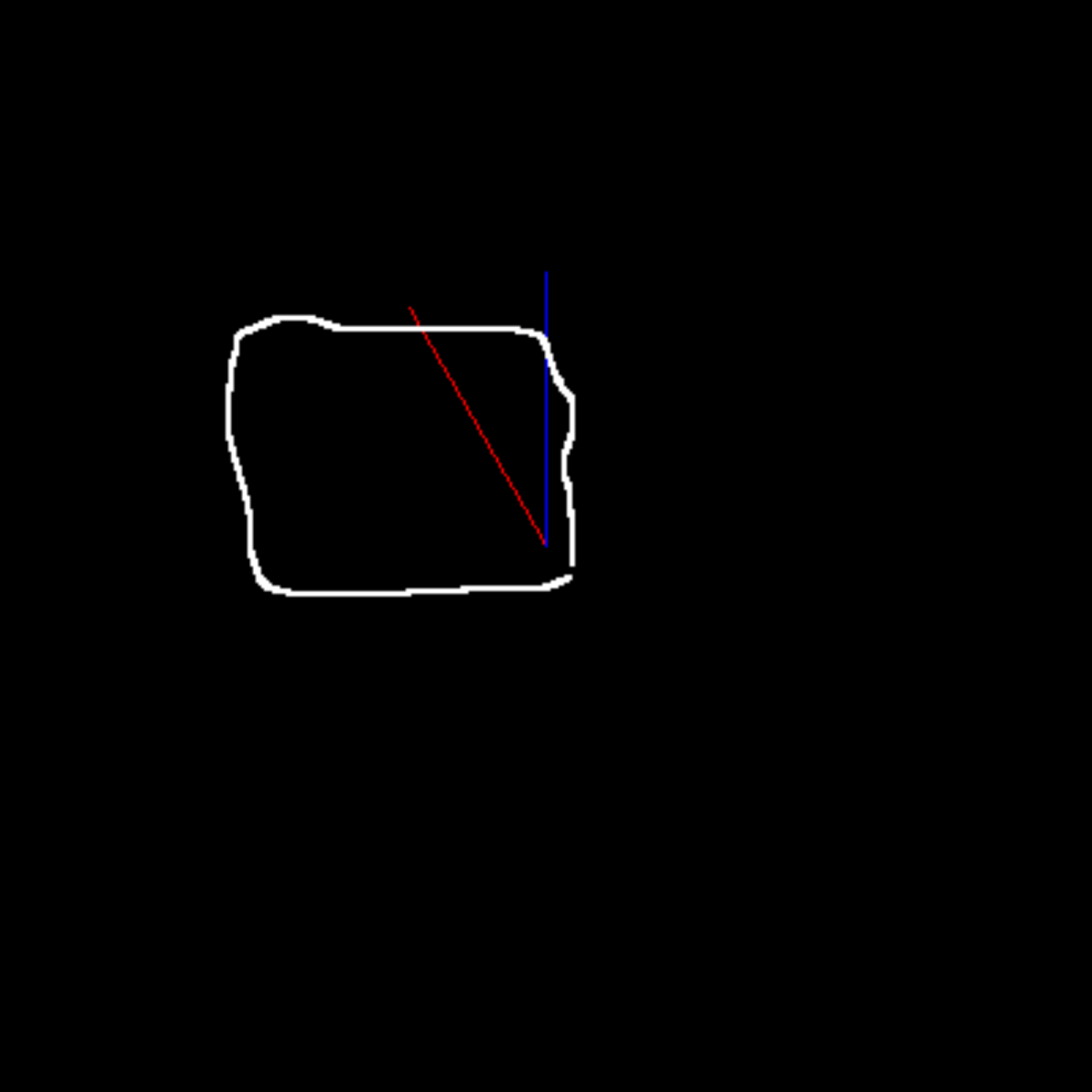} &
\includegraphics[width=0.11\textheight,height=0.11\textheight,trim={0 2cm 3cm 0},clip]{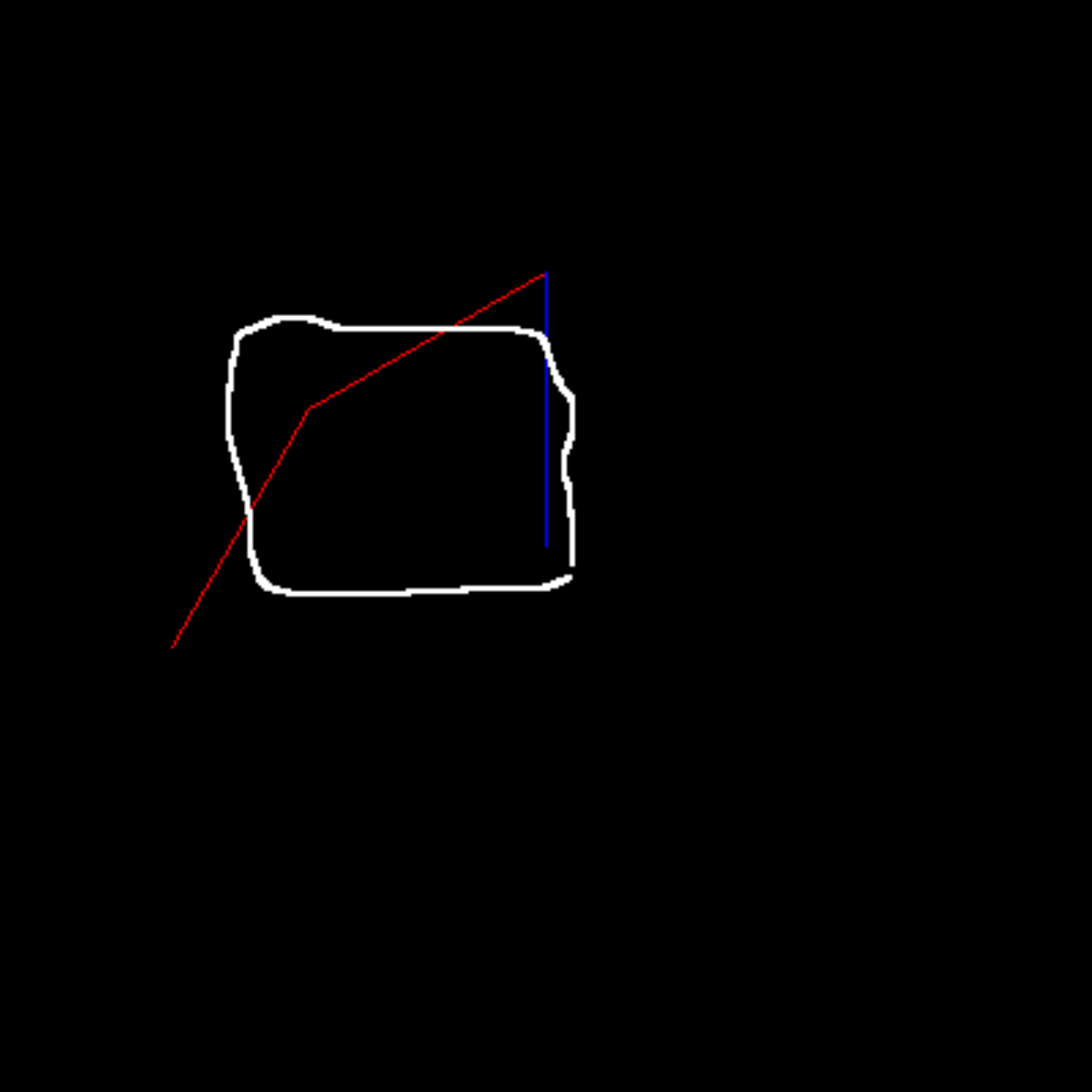} \\
\includegraphics[width=0.11\textheight,height=0.11\textheight,trim={0 2cm 3cm 0},clip]{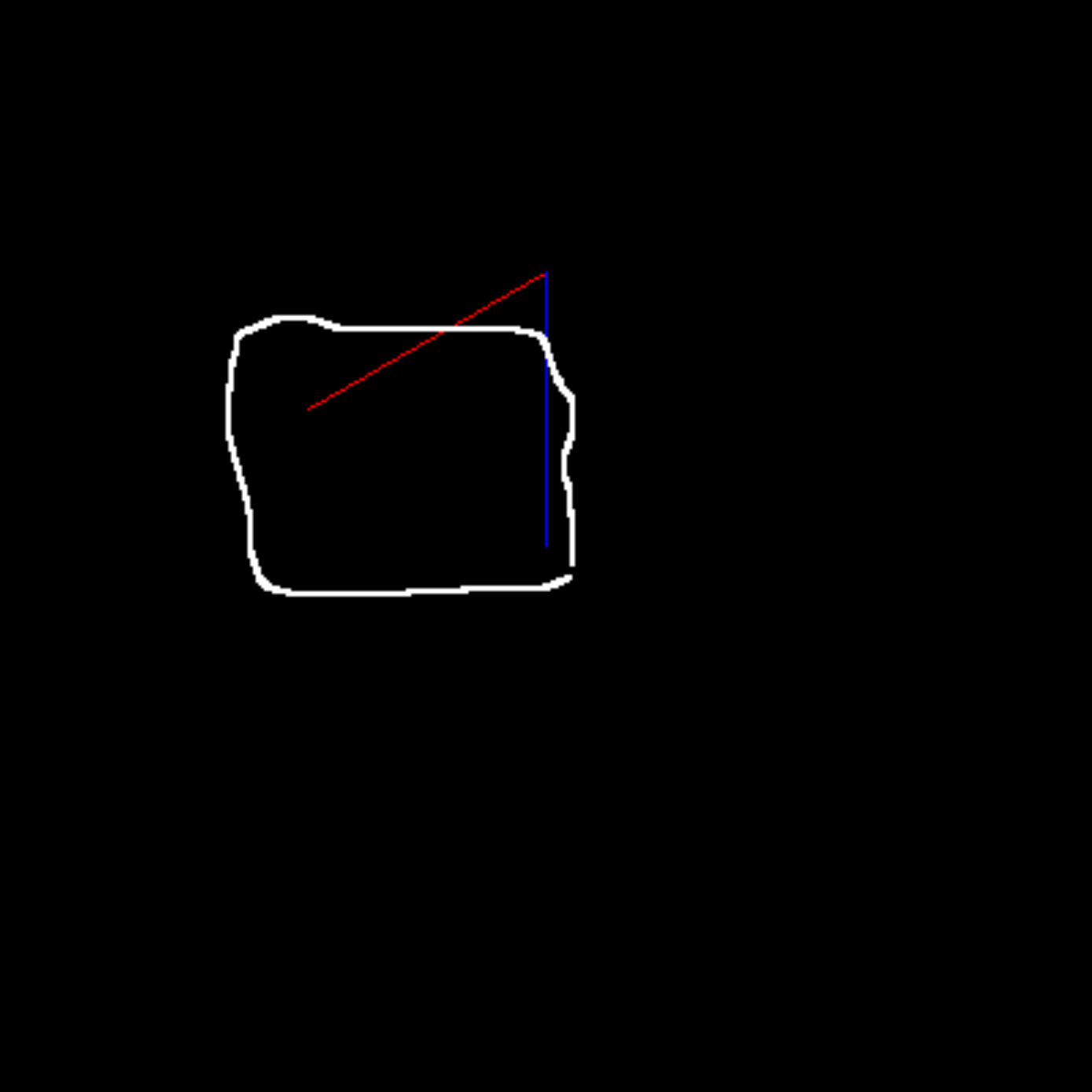} &
\includegraphics[width=0.11\textheight,height=0.11\textheight,trim={0 2cm 3cm 0},clip]{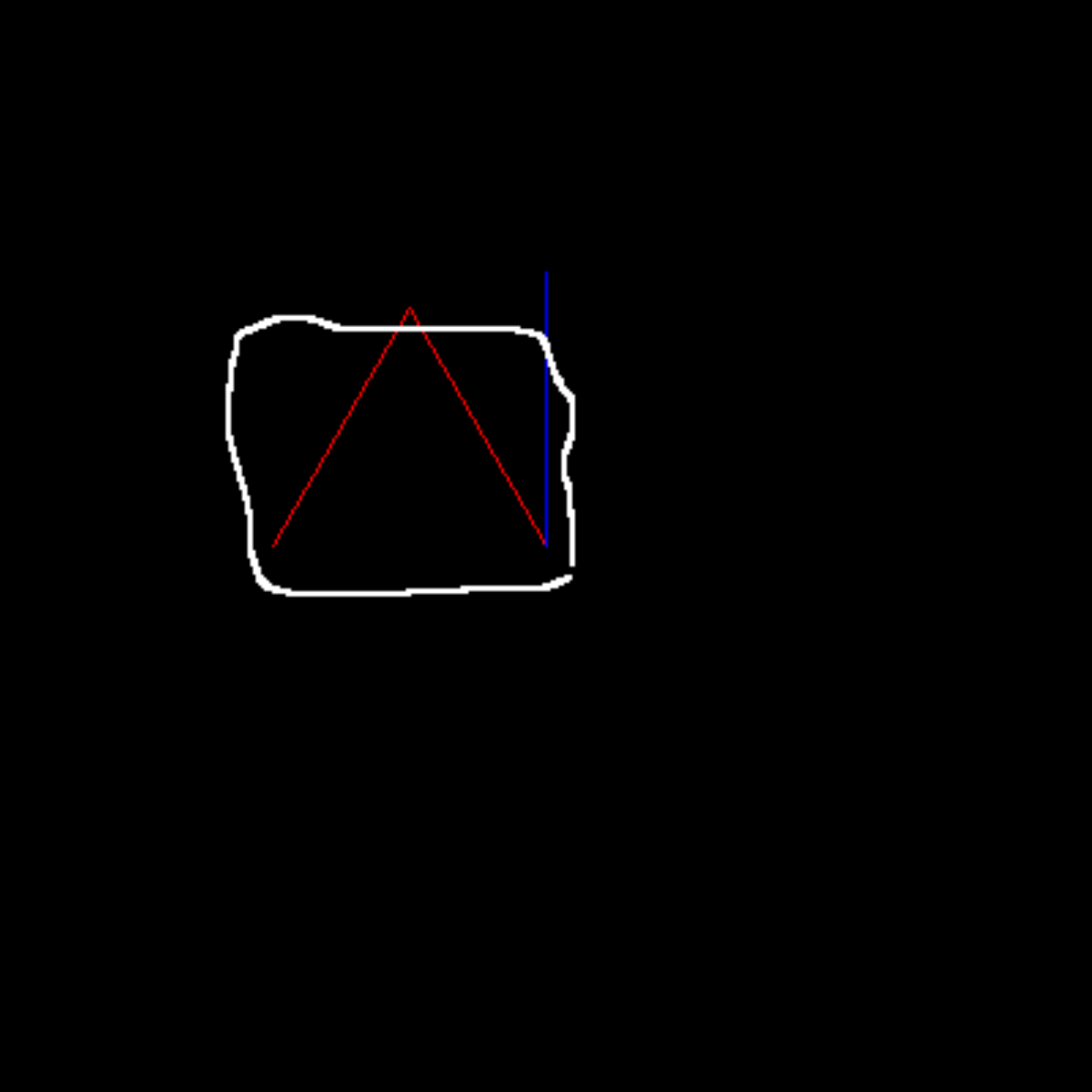} &
\includegraphics[width=0.11\textheight,height=0.11\textheight,trim={0 2cm 3cm 0},clip]{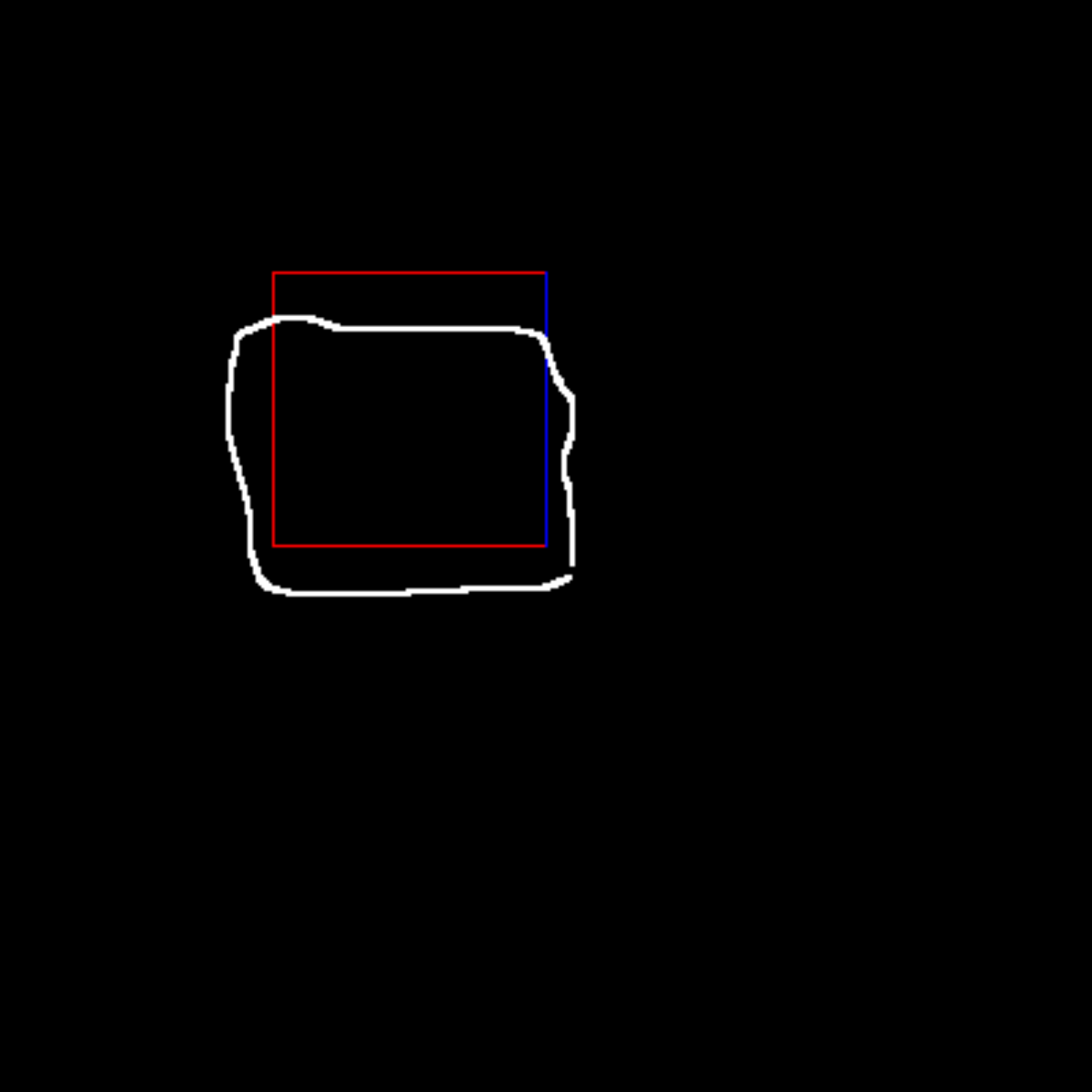}
\end{tabular}
\end{center}
\caption{
	A noisy trajectory (white) atop improving completions $\hat{p}_1, \hat{p}_2, \dots, \hat{p}_6$ (red)
	and the initial program $p$ (blue).
}
\label{fig:searchseq}
\end{figure}

Our first algorithm is a variant of iterative deepening, which we call iterative
deepening program search (IDPS). We use a path-checking depth-limited search as
a subroutine to minimize space requirements. Unlike traditional AI search
procedures, IDPS returns not one but a sequence of programs
$\hat{p}_1,\hat{p}_2,\dots$ whose Hausdorff distance from the target trajectory
$t^*$ strictly decreases. At iteration $d$, we initialize $\epsilon$ to
$d_H(I(p_0),t^*)$ and expand the search graph up to depth $d$. Then, we iterate
through all depth-$d$ programs $p$, checking if $d_H(I(p), t^*) \le \epsilon$.
When such a program is found, we emit it to our output sequence, and update
$\epsilon$ to $d_H(I(p), t^*)$ so that future goal states must be strictly
better than $p$.

\textbf{Figure}~\ref{fig:searchseq} displays an example of such a sequence,
where $t^*$ is a noisy square and the initial program draws only a line. By convention,
we include the initial program $p_0$ in the sequence.

\subsection{Sampling search}

Our second algorithm is a sample-based search shown in
\textbf{Algorithm}~\ref{alg:samplesearch}. Here, we use a corpus of user programs and
trajectories to guide our search. \textbf{Algorithm}~\ref{alg:samplesearch} can be
described as searching from a graph rooted at $p_0$. The initial program $p_0$
is represented by a sequence of editing commands. We use the notation $\bar{p} = p:c$ to
represent the program $\bar{p}$ resulting in appending command $c$ to the end of
command list $p$. The algorithm samples commands $c$ from a distribution
$\Pr_{c|p}$ parameterized by command sequence $p$. A budget of $b$ candidates are
sampled in total. Each of the $\frac{b}{C}$ sampling rounds begins at $p_0$
and samples $C$ commands, sequentially appending and evaluating them. The
candidate minimizing $d_H(I(\cdot),t^*)$ among all samples candidates is returned.

\begin{algorithm}
  \begin{algorithmic}
    \caption{Sampling Search}\label{alg:samplesearch}
    \Require Budget $b$, cost $C$, initial program $p_0$, visual specification
    $t^*$
    \Ensure Program $best$
    \Procedure{SamplingSearch}{$b, C, p_0, t^*$}
      \State $best \gets p_0$ 
      \For{each $i$ in $1 \dots \frac{b}{C}$}
      \For{each $j$ in $1 \dots C$}
      \State{$c_j \sim \Pr(c|p_{j-1})$}
      \State{$p_j \gets p_{j-1} : c_j$} 
      \If{$d_H(I(p_j), t^*) < d_H(I(best), t^*)$}
      \State $best \gets p_j$
      \EndIf
      \EndFor
      \EndFor
      \Return{$best$}
    \EndProcedure
  \end{algorithmic}
\end{algorithm}

What distinguishes variants of this algorithm is how the distribution $\Pr(c|p)$ is
modeled. We factor $\Pr(c|p)$ this into a bigram model over command types and a distribution over
command arguments. That is, we map the command sequence $p = \{ c_i \}_{i=1}^{|p|}$
to a coarsened sequence $\{ \tilde{c}_i \}_{i=1}^{|p|}$ by discarding arguments
so that $\tilde{c}_i \in$ \{\textrm{Get}, \textrm{Remove}, \textrm{Connect}, \textrm{Change}, \textrm{Separate}\}.
We then draw $\tilde{c}_{|p|+1}$ from $\Pr(\cdot|\tilde{c}_{|p|})$, a
Markov chain over the coarsened tag sequence.

To sample $c_{|p|+1}$ from $\tilde{c}_{|p|+1}$, we have two models: 
a \textbf{uniform} model and a \textbf{non-uniform} model. On the uniform model,
we uniformly sample the $\{\textrm{Type}\}$, $\{BlockId\}$, $\{\textrm{Val}\}$
arguments from the command language from all available types, values, and -- in the
case $\{BlockId\}$ -- all available positions in the current program.

For the non-uniform model, we make the following observation about the
process of editing programs: locations to modify the current program are chosen
with a particular focus in mind. Specifically, when a user is connecting a block
to another, the source block is more often the last block added to the
workspace, while the destination block is often the next-to-last block added to
the workspace. We construct a simple model to accommodate this observation. In
choosing $\{BlockId\}$ arguments, we designate a probability $\lambda_{-1}$ that
the source block is the last block added to the workspace. We assign probability
mass $(1 - \lambda_{-1})$ uniformly over other feasible blocks. Then we sample a
destination block as the next-to-last block with probability $\lambda_{-2}$,
reserving probability $(1 - \lambda_{-2})$ to be uniformly distributed over all
remaining feasible choices.

All of the these probabilities in the above models can be estimated from our
corpus. We estimate the transition probabilities $\Pr(\tilde{c_j}|\tilde{c}_{j-1})$ over the
coarsened command sequences smoothing with pseudo-counts of 1. We estimate
$\lambda_{-1}$ and $\lambda_{-2}$ by taking the empirical proportion of such
decisions over the corpus. The product of the distribution over command types
and the distribution over command arguments defines the distribution
$\Pr(c|p_{j-1})$ shown in \textbf{Algorithm}~\ref{alg:samplesearch}.

\subsection{A Computational Speedup}

While the Hausdorff distance $d_H$, used in both IDPS and sampling search
algorithms, is a natural choice for scoring the quality of a fit, the
computation of $d_H(X,Y)$ becomes prohibitive, as it is quadratic in the number
of points of the two point sets $x$ and $y$ to be compared. Let us say we have
some threshold $\alpha > 0$ and we wish to determine if $d_H(X,Y) < \alpha$. In
many cases where $d_H(X,Y) \ge \alpha$, we may avoid some of this computation.
If there exists a point $x \in X$ such that $\forall y \in Y, d(x,y) \ge
\alpha$, then we may terminate our computation, as $d_H(X,Y) \ge \alpha$.
Furthermore, if for any $x \in X$, we can find some $y \in Y$ such that $d(x,y)
< \alpha$, we may omit computation of any further $d(x,y')$ for $y' \in Y$,
because $\min_{y \in Y} d(x,y) < \alpha$. These speed ups, however, avoid
computation only if $d_H \ge \alpha$. They do not compute $d_H$. Therefore, these
speed ups are employed in the algorithms above by replacing any inequality
condition involving $d_H$. If the $d_H < \alpha$, the full quadratic computation
of $d_H$ is performed to compute its value.

\section{Evaluation}

\begin{figure}[htb]
\begin{center}
\begin{tabular}{ccc}
\includegraphics[width=0.14\textwidth]{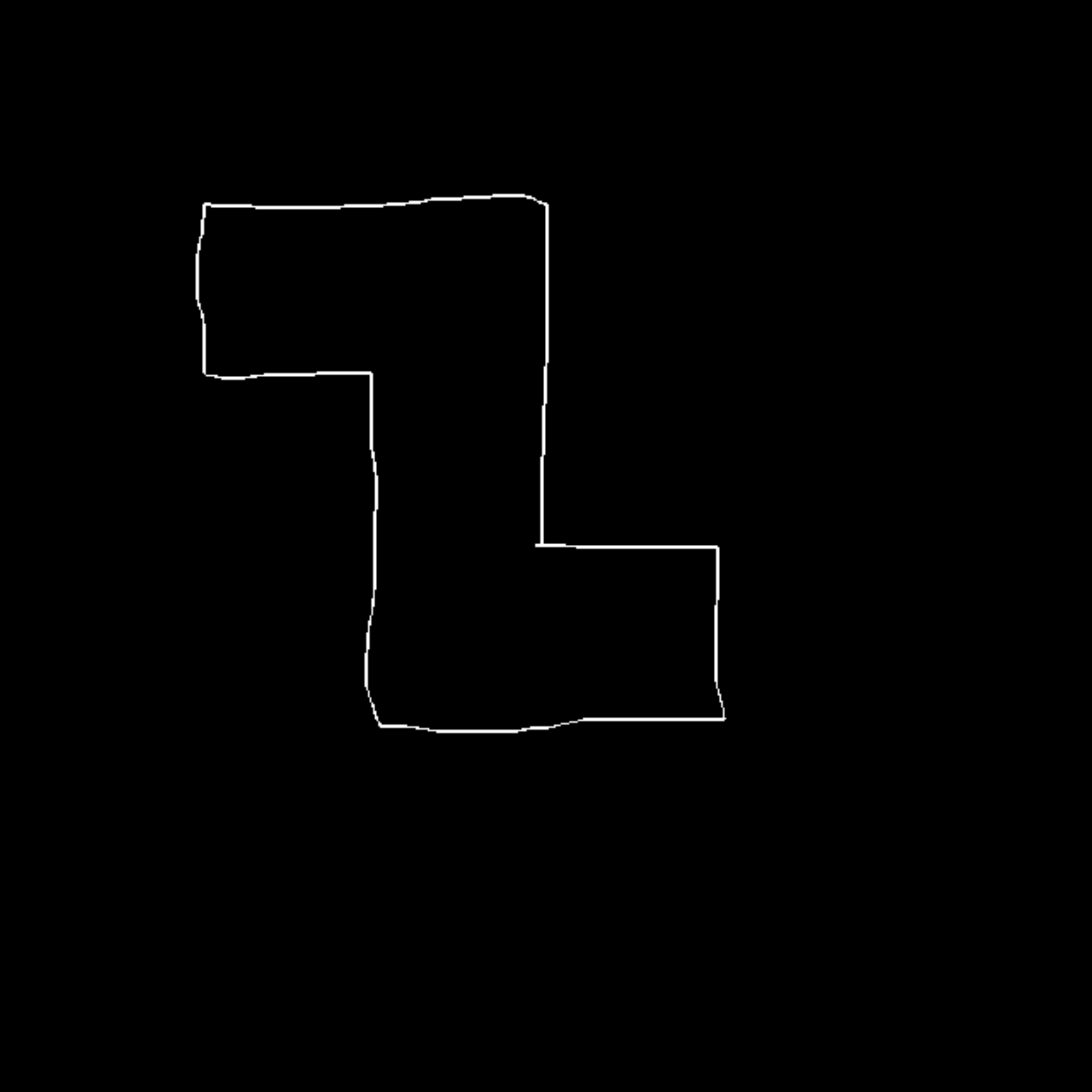} &
\includegraphics[width=0.14\textwidth]{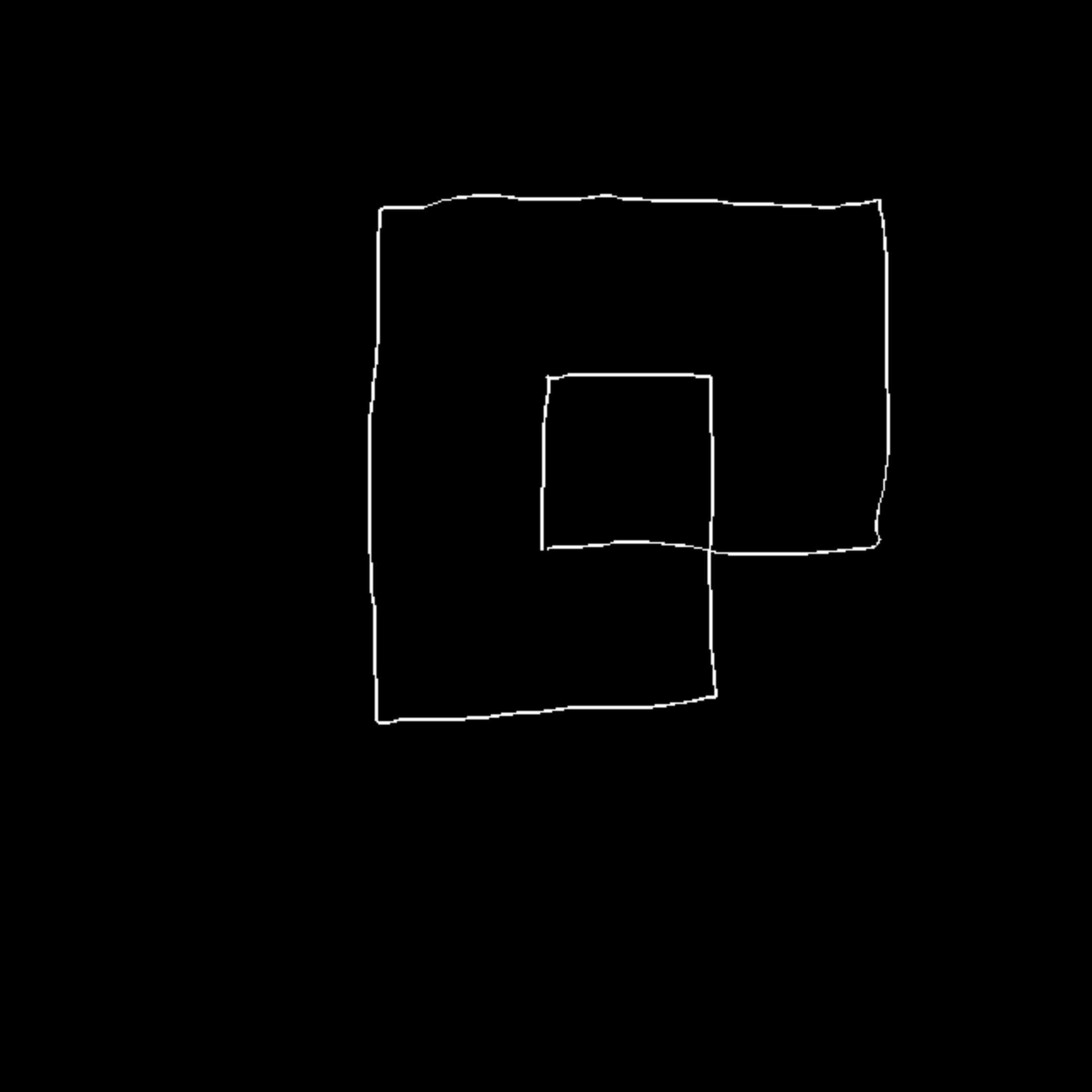} &
\includegraphics[width=0.14\textwidth]{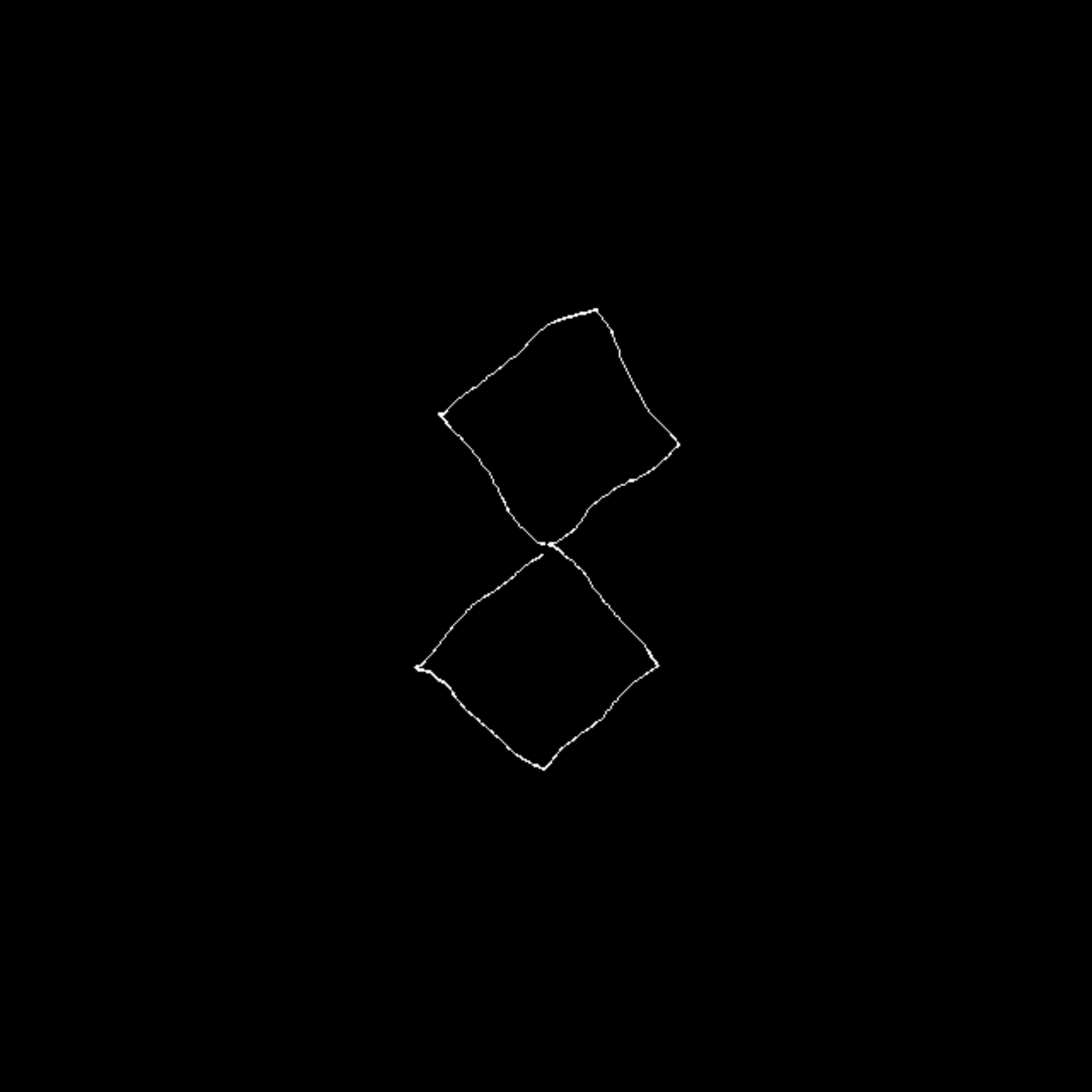} \\
\includegraphics[width=0.14\textwidth]{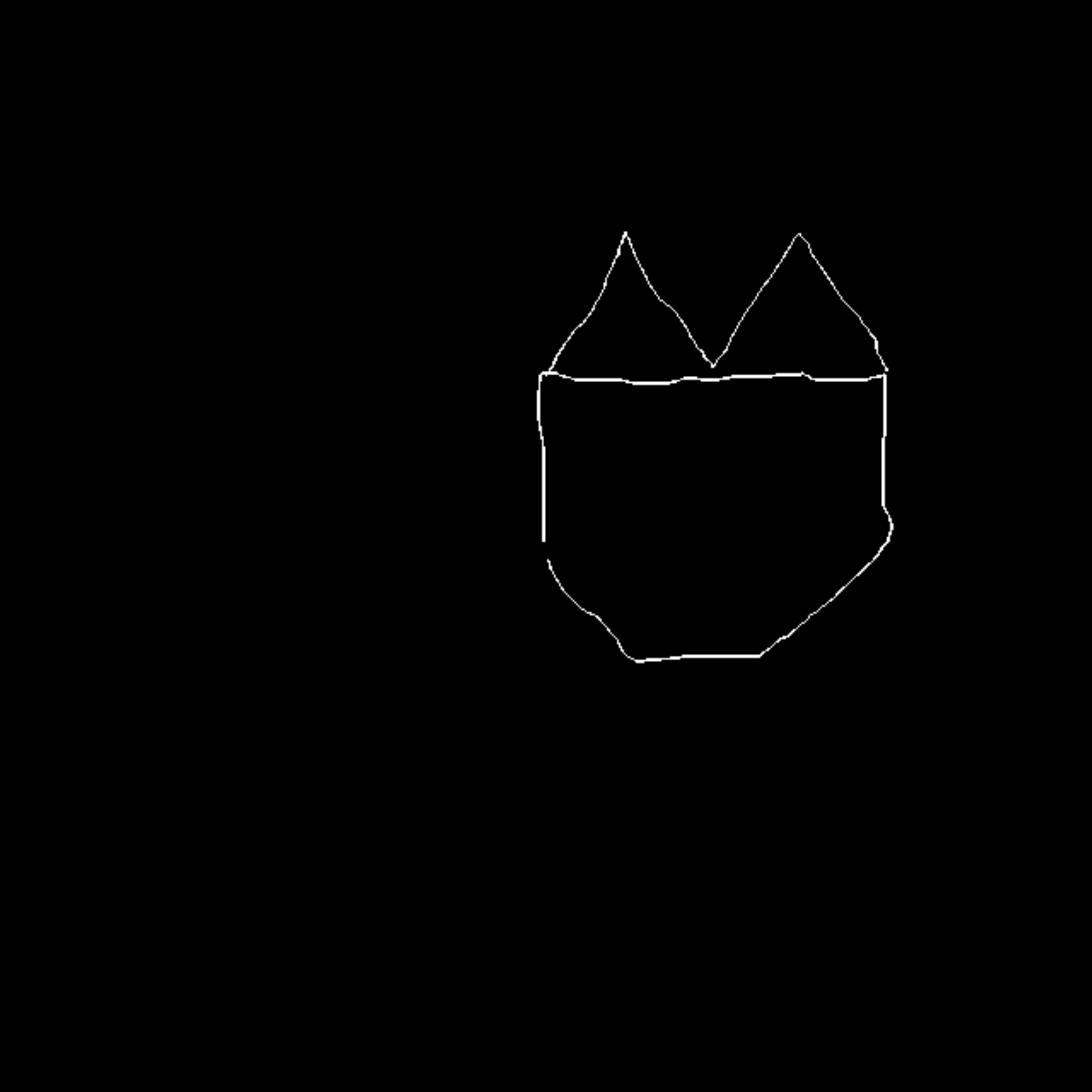} &
\includegraphics[width=0.14\textwidth]{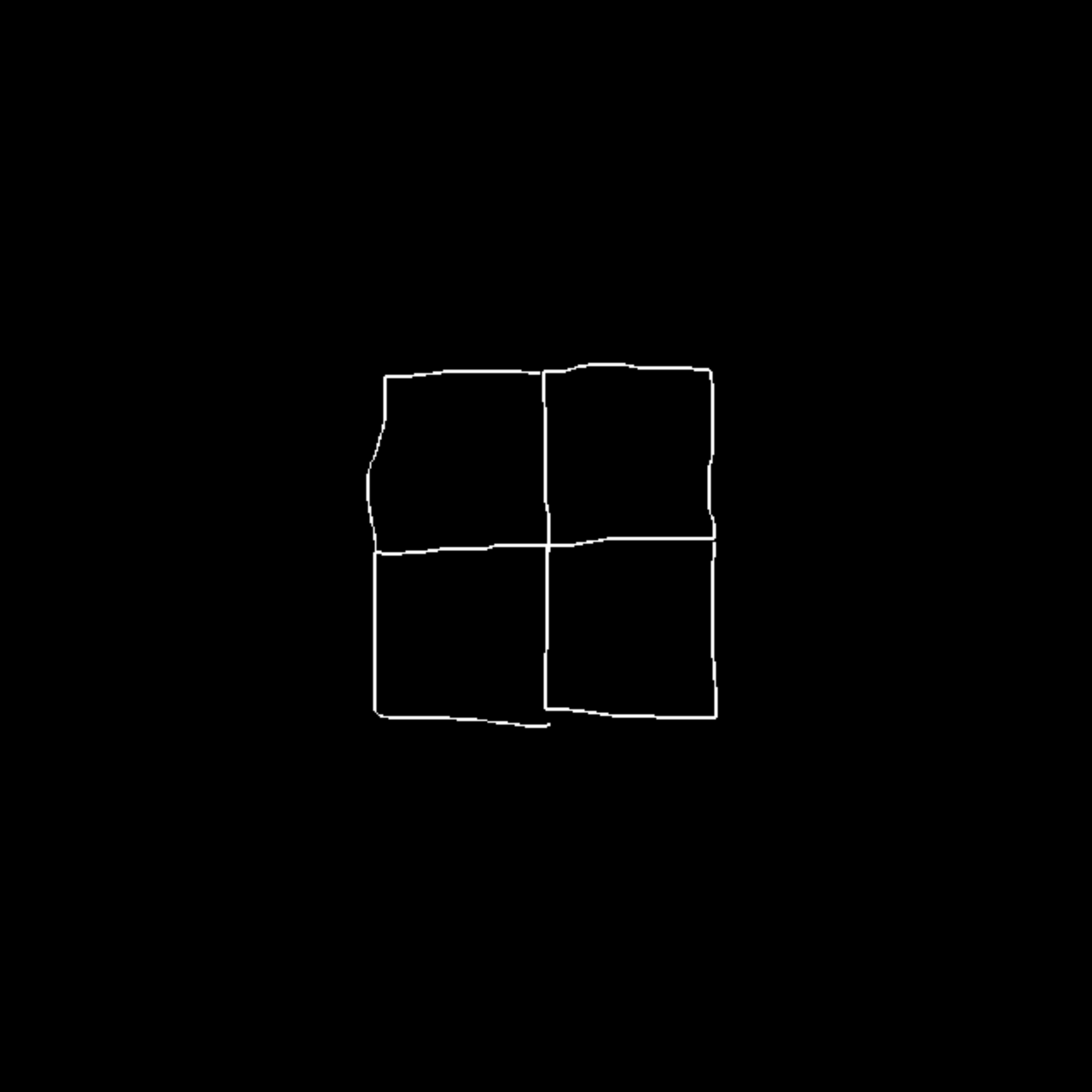} &
\includegraphics[width=0.14\textwidth]{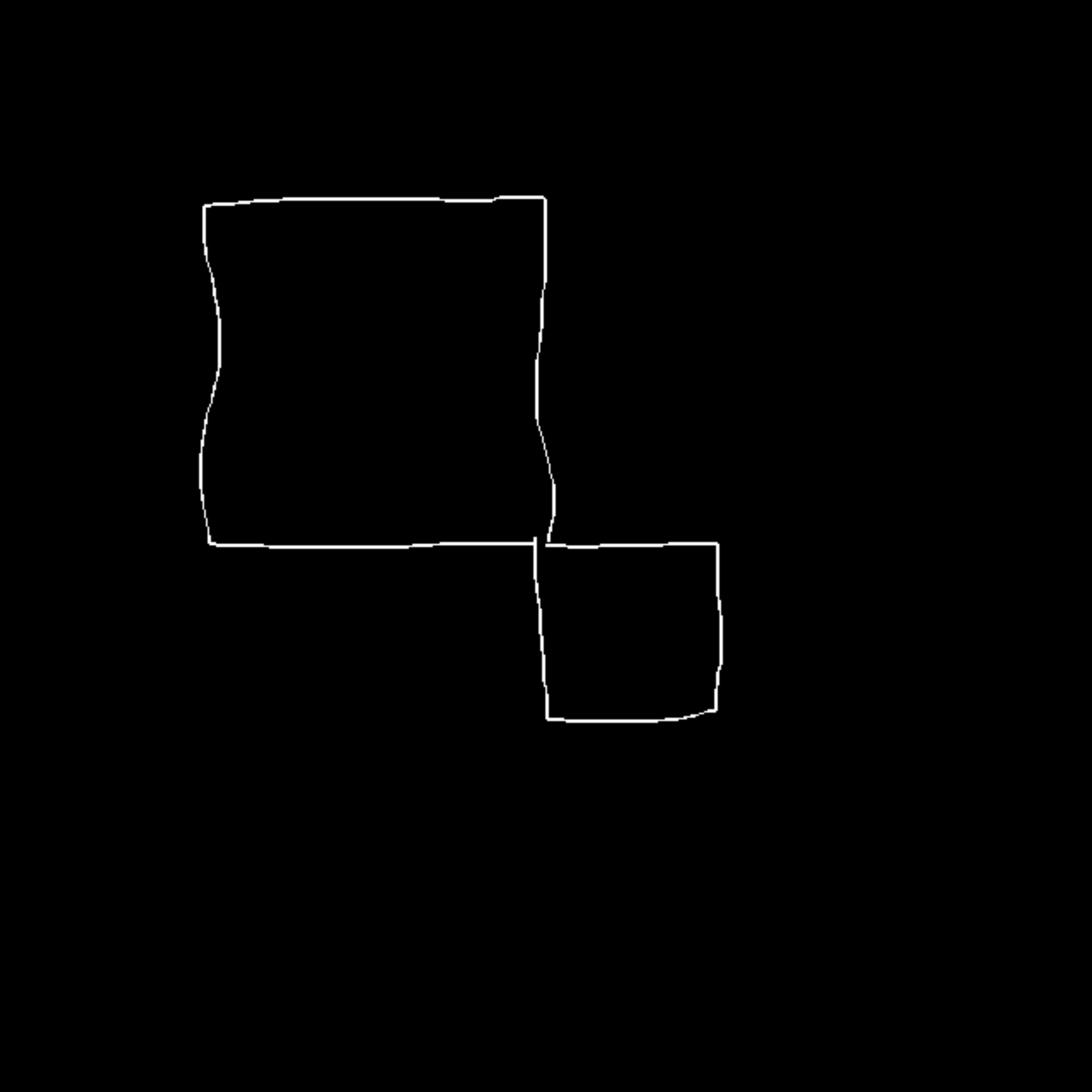}
\end{tabular}
\end{center}
\caption{Sample trajectories drawn by participants.}
\label{fig:sample}
\end{figure}

We solicit a corpus of $\testsetsize = 23$ programs and their visual specifications from
11 volunteer study participants, who range in programming experience from novice to
professional. \textbf{Figure}~\ref{fig:corpus} and \textbf{Figure}~\ref{fig:sample} give an examples
of a participant program and some participant specifications, respectively.
\begin{figure}[t]
  \begin{itemize}
    \setlength\itemsep{0.1em}
  \item Get a repeat block
  \item Get a turn block
  \item Connect block 2 inside block 1
  \item Change 30 in block 2 to 270
  \item Get a move block
  \item Connect block 3 under block 2
  \item Get a repeat block
  \\ \dots
\end{itemize}
\caption{Fragment of a program from the corpus.}
\label{fig:corpus}
\end{figure}
To construct this corpus, we employ the following data collection procedure:

\textbf{Step 1} Each participant is educated in the capabilities of our turtle language
and the editor environment by completing an introductory set of exercises.

\textbf{Step 2} The participant is instructed to draw a trajectory on a standard
canvas. Let $\testt^{(i)}$ represent this visual specification of the intended of
the program, as drawn by participant $i$.

\textbf{Step 3} The participant is instructed to compose a program in
the turtle language which follows the drawn trajectory as closely as possible.
Let $\testp^{(i)}$ represent the matching program from participant $i$.

We record each step in the participant's programming process as a formal editing
command, c.f. Section~\ref{sec:editing}. We represent the complete program
$\testp^{(i)}$ as the sequence of editing commands
$\{\testc^{(i)}_1,\dots,\testc^{(i)}_{|\testp^{(i)}|}\}$ which produce
$\testp^{(i)}$ if performed in the editor begining at an empty
workspace.

Now consider each of these programs with the final $k$ commands
removed, that is $\testp_{-k}^{(i)} = \{\testc^{(i)}_1,\dots,\testc^{(i)}_{|\testp^{(i)}|-k}\}$. Letting $\alg$ denote
the search algorithm at hand, our interest is in evaluating the performance of $\hat{\testp}^{(i)} =
\alg(\testp_{-k}^{(i)},\testt^{(i)})$, the $k$-ahead performance of our search algorithm.

\begin{table}[b]
\begin{center}
\def\arraystretch{2}%
\begin{tabular}{p{22mm} | l | l} \hline
\textbf{Metric} & \textbf{Notation} & \textbf{Equation} \\ \hline

Accuracy & $Acc_k(p^{(i)},t^{(i)})$ & $ \Ind[\hat{\testp}^{(i)} \in SemEq(p^{(i)})]$ \\

Hausdorff Error & $Err_k(p^{(i)}, t^{(i)})$ & $d_H(t^{(i)},I(\hat{\testp}^{(i)}))$ \\

Relative Error Reduction & $\Delta_k(p^{(i)},t^{(i)})$ & $\frac{Err_k(p_{-k}^{(i)}, t^{(i)}) - Err_k(\hat{\testp}^{(i)}, t^{(i)})}{Err_k(p_{-k}^{(i)}, t^{(i)})}$ \\
  \hline

\end{tabular}
\end{center}
\caption{$k$-ahead metrics}\label{table:metrics}
\end{table}

As our search procedure does not discriminate between syntactically differing programs which produce the
same trajectory (share the same semantics), we consider programs up to semantic equivalence. 
We define the semantic equivalence class of $p \in \programspace$ as
$SemEq(p) = \{ p' \in \programspace : I(p) = I(p') \}$. We then define our $k$-ahead accuracy in
terms of our procedure achieving any semantically equivalent program to the target.

Unfortunately, if we only consider $\hat{p}^{(i)}$ correct when $\hat{p}^{(i)} \in SemEq(p)$, we overlook the
important case where $I(\hat{\testp}^{(i)})$ is a better fit for $t^{(i)}$ than $I(\testp^{(i)})$. This may happen,
for example, if the participant has little programming experience and writes $\testp_i$ incorrectly.
A softer measure of accuracy is simply the Hausdorff distance between $I(\hat{p}^{(i)})$ and the user's trajectory,
though this does not consider scaling. Still softer is the relative reduction in Hausdorff distance from $\testp_{-k}^{(i)}$
to $\hat{\testp}^{(i)}$. \textbf{Table}~\ref{table:metrics} summarizes these $k$-ahead metrics.

We evaluate our method against all metrics. We also report runtime.

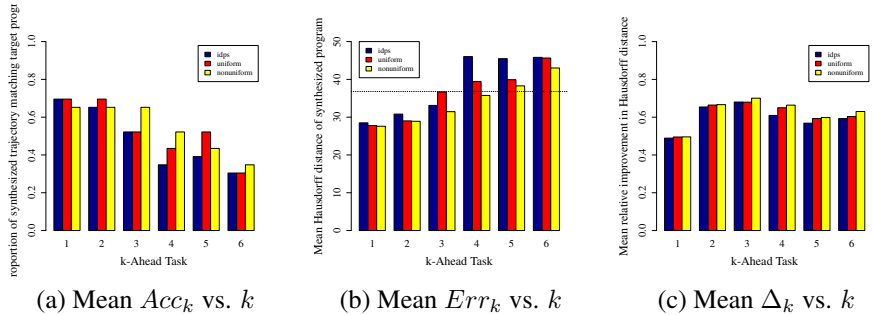
\begin{figure*}[t]
\begin{center} 
\begin{tabular}{ccc}
\resizebox{0.30\textwidth}{!}{
\begin{tikzpicture}[x=1pt,y=1pt]
\definecolor{fillColor}{RGB}{255,255,255}
\path[use as bounding box,fill=fillColor,fill opacity=0.00] (0,0) rectangle (505.89,505.89);
\begin{scope}
\path[clip] (  0.00,  0.00) rectangle (505.89,505.89);
\definecolor{drawColor}{RGB}{0,0,0}
\definecolor{fillColor}{RGB}{0,0,139}

\path[draw=drawColor,line width= 0.4pt,line join=round,line cap=round,fill=fillColor] ( 83.76, 85.68) rectangle ( 99.93,330.08);
\definecolor{fillColor}{RGB}{255,0,0}

\path[draw=drawColor,line width= 0.4pt,line join=round,line cap=round,fill=fillColor] ( 99.93, 85.68) rectangle (116.10,330.08);
\definecolor{fillColor}{RGB}{255,255,0}

\path[draw=drawColor,line width= 0.4pt,line join=round,line cap=round,fill=fillColor] (116.10, 85.68) rectangle (132.28,314.81);
\definecolor{fillColor}{RGB}{0,0,139}

\path[draw=drawColor,line width= 0.4pt,line join=round,line cap=round,fill=fillColor] (148.45, 85.68) rectangle (164.62,314.81);
\definecolor{fillColor}{RGB}{255,0,0}

\path[draw=drawColor,line width= 0.4pt,line join=round,line cap=round,fill=fillColor] (164.62, 85.68) rectangle (180.80,330.08);
\definecolor{fillColor}{RGB}{255,255,0}

\path[draw=drawColor,line width= 0.4pt,line join=round,line cap=round,fill=fillColor] (180.80, 85.68) rectangle (196.97,314.81);
\definecolor{fillColor}{RGB}{0,0,139}

\path[draw=drawColor,line width= 0.4pt,line join=round,line cap=round,fill=fillColor] (213.14, 85.68) rectangle (229.31,268.98);
\definecolor{fillColor}{RGB}{255,0,0}

\path[draw=drawColor,line width= 0.4pt,line join=round,line cap=round,fill=fillColor] (229.31, 85.68) rectangle (245.49,268.98);
\definecolor{fillColor}{RGB}{255,255,0}

\path[draw=drawColor,line width= 0.4pt,line join=round,line cap=round,fill=fillColor] (245.49, 85.68) rectangle (261.66,314.81);
\definecolor{fillColor}{RGB}{0,0,139}

\path[draw=drawColor,line width= 0.4pt,line join=round,line cap=round,fill=fillColor] (277.83, 85.68) rectangle (294.00,207.88);
\definecolor{fillColor}{RGB}{255,0,0}

\path[draw=drawColor,line width= 0.4pt,line join=round,line cap=round,fill=fillColor] (294.00, 85.68) rectangle (310.18,238.43);
\definecolor{fillColor}{RGB}{255,255,0}

\path[draw=drawColor,line width= 0.4pt,line join=round,line cap=round,fill=fillColor] (310.18, 85.68) rectangle (326.35,268.98);
\definecolor{fillColor}{RGB}{0,0,139}

\path[draw=drawColor,line width= 0.4pt,line join=round,line cap=round,fill=fillColor] (342.52, 85.68) rectangle (358.69,223.16);
\definecolor{fillColor}{RGB}{255,0,0}

\path[draw=drawColor,line width= 0.4pt,line join=round,line cap=round,fill=fillColor] (358.69, 85.68) rectangle (374.87,268.98);
\definecolor{fillColor}{RGB}{255,255,0}

\path[draw=drawColor,line width= 0.4pt,line join=round,line cap=round,fill=fillColor] (374.87, 85.68) rectangle (391.04,238.43);
\definecolor{fillColor}{RGB}{0,0,139}

\path[draw=drawColor,line width= 0.4pt,line join=round,line cap=round,fill=fillColor] (407.21, 85.68) rectangle (423.39,192.61);
\definecolor{fillColor}{RGB}{255,0,0}

\path[draw=drawColor,line width= 0.4pt,line join=round,line cap=round,fill=fillColor] (423.39, 85.68) rectangle (439.56,192.61);
\definecolor{fillColor}{RGB}{255,255,0}

\path[draw=drawColor,line width= 0.4pt,line join=round,line cap=round,fill=fillColor] (439.56, 85.68) rectangle (455.73,207.88);
\end{scope}
\begin{scope}
\path[clip] (  0.00,  0.00) rectangle (505.89,505.89);
\definecolor{drawColor}{RGB}{0,0,0}

\node[text=drawColor,anchor=base,inner sep=0pt, outer sep=0pt, scale=  1.96] at (108.02, 55.44) {1};

\node[text=drawColor,anchor=base,inner sep=0pt, outer sep=0pt, scale=  1.96] at (172.71, 55.44) {2};

\node[text=drawColor,anchor=base,inner sep=0pt, outer sep=0pt, scale=  1.96] at (237.40, 55.44) {3};

\node[text=drawColor,anchor=base,inner sep=0pt, outer sep=0pt, scale=  1.96] at (302.09, 55.44) {4};

\node[text=drawColor,anchor=base,inner sep=0pt, outer sep=0pt, scale=  1.96] at (366.78, 55.44) {5};

\node[text=drawColor,anchor=base,inner sep=0pt, outer sep=0pt, scale=  1.96] at (431.47, 55.44) {6};
\end{scope}
\begin{scope}
\path[clip] (  0.00,  0.00) rectangle (505.89,505.89);
\definecolor{drawColor}{RGB}{0,0,0}

\path[draw=drawColor,line width= 0.4pt,line join=round,line cap=round] (351.78,429.78) rectangle (463.38,362.58);
\definecolor{fillColor}{RGB}{0,0,139}

\path[draw=drawColor,line width= 0.4pt,line join=round,line cap=round,fill=fillColor] (364.38,417.18) rectangle (374.46,408.78);
\definecolor{fillColor}{RGB}{255,0,0}

\path[draw=drawColor,line width= 0.4pt,line join=round,line cap=round,fill=fillColor] (364.38,400.38) rectangle (374.46,391.98);
\definecolor{fillColor}{RGB}{255,255,0}

\path[draw=drawColor,line width= 0.4pt,line join=round,line cap=round,fill=fillColor] (364.38,383.58) rectangle (374.46,375.18);

\node[text=drawColor,anchor=base west,inner sep=0pt, outer sep=0pt, scale=  1.40] at (387.06,408.16) {idps};

\node[text=drawColor,anchor=base west,inner sep=0pt, outer sep=0pt, scale=  1.40] at (387.06,391.36) {uniform};

\node[text=drawColor,anchor=base west,inner sep=0pt, outer sep=0pt, scale=  1.40] at (387.06,374.56) {nonuniform};

\node[text=drawColor,anchor=base,inner sep=0pt, outer sep=0pt, scale=  2.24] at (269.74, 21.84) {k-Ahead Task};

\node[text=drawColor,rotate= 90.00,anchor=base,inner sep=0pt, outer sep=0pt, scale=  2.24] at ( 15.12,261.34) {Proportion of synthesized trajectory matching target program};
\end{scope}
\begin{scope}
\path[clip] (  0.00,  0.00) rectangle (505.89,505.89);
\definecolor{drawColor}{RGB}{0,0,0}

\path[draw=drawColor,line width= 0.4pt,line join=round,line cap=round] ( 68.88, 85.68) -- ( 68.88,437.01);

\path[draw=drawColor,line width= 0.4pt,line join=round,line cap=round] ( 68.88, 85.68) -- ( 60.48, 85.68);

\path[draw=drawColor,line width= 0.4pt,line join=round,line cap=round] ( 68.88,155.95) -- ( 60.48,155.95);

\path[draw=drawColor,line width= 0.4pt,line join=round,line cap=round] ( 68.88,226.21) -- ( 60.48,226.21);

\path[draw=drawColor,line width= 0.4pt,line join=round,line cap=round] ( 68.88,296.48) -- ( 60.48,296.48);

\path[draw=drawColor,line width= 0.4pt,line join=round,line cap=round] ( 68.88,366.74) -- ( 60.48,366.74);

\path[draw=drawColor,line width= 0.4pt,line join=round,line cap=round] ( 68.88,437.01) -- ( 60.48,437.01);

\node[text=drawColor,rotate= 90.00,anchor=base,inner sep=0pt, outer sep=0pt, scale=  1.96] at ( 48.72, 85.68) {0.0};

\node[text=drawColor,rotate= 90.00,anchor=base,inner sep=0pt, outer sep=0pt, scale=  1.96] at ( 48.72,155.95) {0.2};

\node[text=drawColor,rotate= 90.00,anchor=base,inner sep=0pt, outer sep=0pt, scale=  1.96] at ( 48.72,226.21) {0.4};

\node[text=drawColor,rotate= 90.00,anchor=base,inner sep=0pt, outer sep=0pt, scale=  1.96] at ( 48.72,296.48) {0.6};

\node[text=drawColor,rotate= 90.00,anchor=base,inner sep=0pt, outer sep=0pt, scale=  1.96] at ( 48.72,366.74) {0.8};

\node[text=drawColor,rotate= 90.00,anchor=base,inner sep=0pt, outer sep=0pt, scale=  1.96] at ( 48.72,437.01) {1.0};
\end{scope}
\end{tikzpicture}} &  
\resizebox{0.30\textwidth}{!}{
\begin{tikzpicture}[x=1pt,y=1pt]
\definecolor{fillColor}{RGB}{255,255,255}
\path[use as bounding box,fill=fillColor,fill opacity=0.00] (0,0) rectangle (505.89,505.89);
\begin{scope}
\path[clip] (  0.00,  0.00) rectangle (505.89,505.89);
\definecolor{drawColor}{RGB}{0,0,0}
\definecolor{fillColor}{RGB}{0,0,139}

\path[draw=drawColor,line width= 0.4pt,line join=round,line cap=round,fill=fillColor] ( 83.76, 85.68) rectangle ( 99.93,285.89);
\definecolor{fillColor}{RGB}{255,0,0}

\path[draw=drawColor,line width= 0.4pt,line join=round,line cap=round,fill=fillColor] ( 99.93, 85.68) rectangle (116.10,280.90);
\definecolor{fillColor}{RGB}{255,255,0}

\path[draw=drawColor,line width= 0.4pt,line join=round,line cap=round,fill=fillColor] (116.10, 85.68) rectangle (132.28,279.58);
\definecolor{fillColor}{RGB}{0,0,139}

\path[draw=drawColor,line width= 0.4pt,line join=round,line cap=round,fill=fillColor] (148.45, 85.68) rectangle (164.62,302.20);
\definecolor{fillColor}{RGB}{255,0,0}

\path[draw=drawColor,line width= 0.4pt,line join=round,line cap=round,fill=fillColor] (164.62, 85.68) rectangle (180.80,289.54);
\definecolor{fillColor}{RGB}{255,255,0}

\path[draw=drawColor,line width= 0.4pt,line join=round,line cap=round,fill=fillColor] (180.80, 85.68) rectangle (196.97,288.83);
\definecolor{fillColor}{RGB}{0,0,139}

\path[draw=drawColor,line width= 0.4pt,line join=round,line cap=round,fill=fillColor] (213.14, 85.68) rectangle (229.31,318.36);
\definecolor{fillColor}{RGB}{255,0,0}

\path[draw=drawColor,line width= 0.4pt,line join=round,line cap=round,fill=fillColor] (229.31, 85.68) rectangle (245.49,343.36);
\definecolor{fillColor}{RGB}{255,255,0}

\path[draw=drawColor,line width= 0.4pt,line join=round,line cap=round,fill=fillColor] (245.49, 85.68) rectangle (261.66,306.72);
\definecolor{fillColor}{RGB}{0,0,139}

\path[draw=drawColor,line width= 0.4pt,line join=round,line cap=round,fill=fillColor] (277.83, 85.68) rectangle (294.00,409.11);
\definecolor{fillColor}{RGB}{255,0,0}

\path[draw=drawColor,line width= 0.4pt,line join=round,line cap=round,fill=fillColor] (294.00, 85.68) rectangle (310.18,362.67);
\definecolor{fillColor}{RGB}{255,255,0}

\path[draw=drawColor,line width= 0.4pt,line join=round,line cap=round,fill=fillColor] (310.18, 85.68) rectangle (326.35,336.91);
\definecolor{fillColor}{RGB}{0,0,139}

\path[draw=drawColor,line width= 0.4pt,line join=round,line cap=round,fill=fillColor] (342.52, 85.68) rectangle (358.69,405.25);
\definecolor{fillColor}{RGB}{255,0,0}

\path[draw=drawColor,line width= 0.4pt,line join=round,line cap=round,fill=fillColor] (358.69, 85.68) rectangle (374.87,366.15);
\definecolor{fillColor}{RGB}{255,255,0}

\path[draw=drawColor,line width= 0.4pt,line join=round,line cap=round,fill=fillColor] (374.87, 85.68) rectangle (391.04,354.68);
\definecolor{fillColor}{RGB}{0,0,139}

\path[draw=drawColor,line width= 0.4pt,line join=round,line cap=round,fill=fillColor] (407.21, 85.68) rectangle (423.39,407.71);
\definecolor{fillColor}{RGB}{255,0,0}

\path[draw=drawColor,line width= 0.4pt,line join=round,line cap=round,fill=fillColor] (423.39, 85.68) rectangle (439.56,406.46);
\definecolor{fillColor}{RGB}{255,255,0}

\path[draw=drawColor,line width= 0.4pt,line join=round,line cap=round,fill=fillColor] (439.56, 85.68) rectangle (455.73,387.96);
\end{scope}
\begin{scope}
\path[clip] (  0.00,  0.00) rectangle (505.89,505.89);
\definecolor{drawColor}{RGB}{0,0,0}

\node[text=drawColor,anchor=base,inner sep=0pt, outer sep=0pt, scale=  1.96] at (108.02, 55.44) {1};

\node[text=drawColor,anchor=base,inner sep=0pt, outer sep=0pt, scale=  1.96] at (172.71, 55.44) {2};

\node[text=drawColor,anchor=base,inner sep=0pt, outer sep=0pt, scale=  1.96] at (237.40, 55.44) {3};

\node[text=drawColor,anchor=base,inner sep=0pt, outer sep=0pt, scale=  1.96] at (302.09, 55.44) {4};

\node[text=drawColor,anchor=base,inner sep=0pt, outer sep=0pt, scale=  1.96] at (366.78, 55.44) {5};

\node[text=drawColor,anchor=base,inner sep=0pt, outer sep=0pt, scale=  1.96] at (431.47, 55.44) {6};
\end{scope}
\begin{scope}
\path[clip] (  0.00,  0.00) rectangle (505.89,505.89);
\definecolor{drawColor}{RGB}{0,0,0}

\node[text=drawColor,anchor=base,inner sep=0pt, outer sep=0pt, scale=  2.24] at (269.74, 21.84) {k-Ahead Task};

\node[text=drawColor,rotate= 90.00,anchor=base,inner sep=0pt, outer sep=0pt, scale=  2.24] at ( 15.12,261.34) {Mean Hausdorff distance of synthesized program};
\end{scope}
\begin{scope}
\path[clip] (  0.00,  0.00) rectangle (505.89,505.89);
\definecolor{drawColor}{RGB}{0,0,0}

\path[draw=drawColor,line width= 0.4pt,line join=round,line cap=round] ( 68.88, 85.68) -- ( 68.88,437.01);

\path[draw=drawColor,line width= 0.4pt,line join=round,line cap=round] ( 68.88, 85.68) -- ( 60.48, 85.68);

\path[draw=drawColor,line width= 0.4pt,line join=round,line cap=round] ( 68.88,155.95) -- ( 60.48,155.95);

\path[draw=drawColor,line width= 0.4pt,line join=round,line cap=round] ( 68.88,226.21) -- ( 60.48,226.21);

\path[draw=drawColor,line width= 0.4pt,line join=round,line cap=round] ( 68.88,296.48) -- ( 60.48,296.48);

\path[draw=drawColor,line width= 0.4pt,line join=round,line cap=round] ( 68.88,366.74) -- ( 60.48,366.74);

\path[draw=drawColor,line width= 0.4pt,line join=round,line cap=round] ( 68.88,437.01) -- ( 60.48,437.01);

\node[text=drawColor,rotate= 90.00,anchor=base,inner sep=0pt, outer sep=0pt, scale=  1.96] at ( 48.72, 85.68) {0};

\node[text=drawColor,rotate= 90.00,anchor=base,inner sep=0pt, outer sep=0pt, scale=  1.96] at ( 48.72,155.95) {10};

\node[text=drawColor,rotate= 90.00,anchor=base,inner sep=0pt, outer sep=0pt, scale=  1.96] at ( 48.72,226.21) {20};

\node[text=drawColor,rotate= 90.00,anchor=base,inner sep=0pt, outer sep=0pt, scale=  1.96] at ( 48.72,296.48) {30};

\node[text=drawColor,rotate= 90.00,anchor=base,inner sep=0pt, outer sep=0pt, scale=  1.96] at ( 48.72,366.74) {40};

\node[text=drawColor,rotate= 90.00,anchor=base,inner sep=0pt, outer sep=0pt, scale=  1.96] at ( 48.72,437.01) {50};
\end{scope}
\begin{scope}
\path[clip] ( 68.88, 85.68) rectangle (470.61,437.01);
\definecolor{drawColor}{RGB}{0,0,0}

\path[draw=drawColor,line width= 0.4pt,line join=round,line cap=round] ( 83.76,437.01) rectangle (195.36,369.81);
\definecolor{fillColor}{RGB}{0,0,139}

\path[draw=drawColor,line width= 0.4pt,line join=round,line cap=round,fill=fillColor] ( 96.36,424.41) rectangle (106.44,416.01);
\definecolor{fillColor}{RGB}{255,0,0}

\path[draw=drawColor,line width= 0.4pt,line join=round,line cap=round,fill=fillColor] ( 96.36,407.61) rectangle (106.44,399.21);
\definecolor{fillColor}{RGB}{255,255,0}

\path[draw=drawColor,line width= 0.4pt,line join=round,line cap=round,fill=fillColor] ( 96.36,390.81) rectangle (106.44,382.41);

\node[text=drawColor,anchor=base west,inner sep=0pt, outer sep=0pt, scale=  1.40] at (119.04,415.39) {idps};

\node[text=drawColor,anchor=base west,inner sep=0pt, outer sep=0pt, scale=  1.40] at (119.04,398.59) {uniform};

\node[text=drawColor,anchor=base west,inner sep=0pt, outer sep=0pt, scale=  1.40] at (119.04,381.79) {nonuniform};

\path[draw=drawColor,line width= 0.4pt,dash pattern=on 1pt off 3pt ,line join=round,line cap=round] ( 68.88,344.24) -- (470.61,344.24);
\end{scope}
\end{tikzpicture}} &
\resizebox{0.30\textwidth}{!}{
\begin{tikzpicture}[x=1pt,y=1pt]
\definecolor{fillColor}{RGB}{255,255,255}
\path[use as bounding box,fill=fillColor,fill opacity=0.00] (0,0) rectangle (505.89,505.89);
\begin{scope}
\path[clip] (  0.00,  0.00) rectangle (505.89,505.89);
\definecolor{drawColor}{RGB}{0,0,0}
\definecolor{fillColor}{RGB}{0,0,139}

\path[draw=drawColor,line width= 0.4pt,line join=round,line cap=round,fill=fillColor] ( 83.76, 85.68) rectangle ( 99.93,257.54);
\definecolor{fillColor}{RGB}{255,0,0}

\path[draw=drawColor,line width= 0.4pt,line join=round,line cap=round,fill=fillColor] ( 99.93, 85.68) rectangle (116.10,259.61);
\definecolor{fillColor}{RGB}{255,255,0}

\path[draw=drawColor,line width= 0.4pt,line join=round,line cap=round,fill=fillColor] (116.10, 85.68) rectangle (132.28,259.83);
\definecolor{fillColor}{RGB}{0,0,139}

\path[draw=drawColor,line width= 0.4pt,line join=round,line cap=round,fill=fillColor] (148.45, 85.68) rectangle (164.62,315.42);
\definecolor{fillColor}{RGB}{255,0,0}

\path[draw=drawColor,line width= 0.4pt,line join=round,line cap=round,fill=fillColor] (164.62, 85.68) rectangle (180.80,318.99);
\definecolor{fillColor}{RGB}{255,255,0}

\path[draw=drawColor,line width= 0.4pt,line join=round,line cap=round,fill=fillColor] (180.80, 85.68) rectangle (196.97,319.79);
\definecolor{fillColor}{RGB}{0,0,139}

\path[draw=drawColor,line width= 0.4pt,line join=round,line cap=round,fill=fillColor] (213.14, 85.68) rectangle (229.31,324.63);
\definecolor{fillColor}{RGB}{255,0,0}

\path[draw=drawColor,line width= 0.4pt,line join=round,line cap=round,fill=fillColor] (229.31, 85.68) rectangle (245.49,324.36);
\definecolor{fillColor}{RGB}{255,255,0}

\path[draw=drawColor,line width= 0.4pt,line join=round,line cap=round,fill=fillColor] (245.49, 85.68) rectangle (261.66,331.79);
\definecolor{fillColor}{RGB}{0,0,139}

\path[draw=drawColor,line width= 0.4pt,line join=round,line cap=round,fill=fillColor] (277.83, 85.68) rectangle (294.00,299.72);
\definecolor{fillColor}{RGB}{255,0,0}

\path[draw=drawColor,line width= 0.4pt,line join=round,line cap=round,fill=fillColor] (294.00, 85.68) rectangle (310.18,313.98);
\definecolor{fillColor}{RGB}{255,255,0}

\path[draw=drawColor,line width= 0.4pt,line join=round,line cap=round,fill=fillColor] (310.18, 85.68) rectangle (326.35,318.86);
\definecolor{fillColor}{RGB}{0,0,139}

\path[draw=drawColor,line width= 0.4pt,line join=round,line cap=round,fill=fillColor] (342.52, 85.68) rectangle (358.69,285.33);
\definecolor{fillColor}{RGB}{255,0,0}

\path[draw=drawColor,line width= 0.4pt,line join=round,line cap=round,fill=fillColor] (358.69, 85.68) rectangle (374.87,293.93);
\definecolor{fillColor}{RGB}{255,255,0}

\path[draw=drawColor,line width= 0.4pt,line join=round,line cap=round,fill=fillColor] (374.87, 85.68) rectangle (391.04,295.81);
\definecolor{fillColor}{RGB}{0,0,139}

\path[draw=drawColor,line width= 0.4pt,line join=round,line cap=round,fill=fillColor] (407.21, 85.68) rectangle (423.39,293.89);
\definecolor{fillColor}{RGB}{255,0,0}

\path[draw=drawColor,line width= 0.4pt,line join=round,line cap=round,fill=fillColor] (423.39, 85.68) rectangle (439.56,297.74);
\definecolor{fillColor}{RGB}{255,255,0}

\path[draw=drawColor,line width= 0.4pt,line join=round,line cap=round,fill=fillColor] (439.56, 85.68) rectangle (455.73,307.06);
\end{scope}
\begin{scope}
\path[clip] (  0.00,  0.00) rectangle (505.89,505.89);
\definecolor{drawColor}{RGB}{0,0,0}

\node[text=drawColor,anchor=base,inner sep=0pt, outer sep=0pt, scale=  1.96] at (108.02, 55.44) {1};

\node[text=drawColor,anchor=base,inner sep=0pt, outer sep=0pt, scale=  1.96] at (172.71, 55.44) {2};

\node[text=drawColor,anchor=base,inner sep=0pt, outer sep=0pt, scale=  1.96] at (237.40, 55.44) {3};

\node[text=drawColor,anchor=base,inner sep=0pt, outer sep=0pt, scale=  1.96] at (302.09, 55.44) {4};

\node[text=drawColor,anchor=base,inner sep=0pt, outer sep=0pt, scale=  1.96] at (366.78, 55.44) {5};

\node[text=drawColor,anchor=base,inner sep=0pt, outer sep=0pt, scale=  1.96] at (431.47, 55.44) {6};
\end{scope}
\begin{scope}
\path[clip] (  0.00,  0.00) rectangle (505.89,505.89);
\definecolor{drawColor}{RGB}{0,0,0}

\path[draw=drawColor,line width= 0.4pt,line join=round,line cap=round] (351.78,429.78) rectangle (463.38,362.58);
\definecolor{fillColor}{RGB}{0,0,139}

\path[draw=drawColor,line width= 0.4pt,line join=round,line cap=round,fill=fillColor] (364.38,417.18) rectangle (374.46,408.78);
\definecolor{fillColor}{RGB}{255,0,0}

\path[draw=drawColor,line width= 0.4pt,line join=round,line cap=round,fill=fillColor] (364.38,400.38) rectangle (374.46,391.98);
\definecolor{fillColor}{RGB}{255,255,0}

\path[draw=drawColor,line width= 0.4pt,line join=round,line cap=round,fill=fillColor] (364.38,383.58) rectangle (374.46,375.18);

\node[text=drawColor,anchor=base west,inner sep=0pt, outer sep=0pt, scale=  1.40] at (387.06,408.16) {idps};

\node[text=drawColor,anchor=base west,inner sep=0pt, outer sep=0pt, scale=  1.40] at (387.06,391.36) {uniform};

\node[text=drawColor,anchor=base west,inner sep=0pt, outer sep=0pt, scale=  1.40] at (387.06,374.56) {nonuniform};

\node[text=drawColor,anchor=base,inner sep=0pt, outer sep=0pt, scale=  2.24] at (269.74, 21.84) {k-Ahead Task};

\node[text=drawColor,rotate= 90.00,anchor=base,inner sep=0pt, outer sep=0pt, scale=  2.24] at ( 15.12,261.34) {Mean relative improvement in Hausdorff distance};
\end{scope}
\begin{scope}
\path[clip] (  0.00,  0.00) rectangle (505.89,505.89);
\definecolor{drawColor}{RGB}{0,0,0}

\path[draw=drawColor,line width= 0.4pt,line join=round,line cap=round] ( 68.88, 85.68) -- ( 68.88,437.01);

\path[draw=drawColor,line width= 0.4pt,line join=round,line cap=round] ( 68.88, 85.68) -- ( 60.48, 85.68);

\path[draw=drawColor,line width= 0.4pt,line join=round,line cap=round] ( 68.88,155.95) -- ( 60.48,155.95);

\path[draw=drawColor,line width= 0.4pt,line join=round,line cap=round] ( 68.88,226.21) -- ( 60.48,226.21);

\path[draw=drawColor,line width= 0.4pt,line join=round,line cap=round] ( 68.88,296.48) -- ( 60.48,296.48);

\path[draw=drawColor,line width= 0.4pt,line join=round,line cap=round] ( 68.88,366.74) -- ( 60.48,366.74);

\path[draw=drawColor,line width= 0.4pt,line join=round,line cap=round] ( 68.88,437.01) -- ( 60.48,437.01);

\node[text=drawColor,rotate= 90.00,anchor=base,inner sep=0pt, outer sep=0pt, scale=  1.96] at ( 48.72, 85.68) {0.0};

\node[text=drawColor,rotate= 90.00,anchor=base,inner sep=0pt, outer sep=0pt, scale=  1.96] at ( 48.72,155.95) {0.2};

\node[text=drawColor,rotate= 90.00,anchor=base,inner sep=0pt, outer sep=0pt, scale=  1.96] at ( 48.72,226.21) {0.4};

\node[text=drawColor,rotate= 90.00,anchor=base,inner sep=0pt, outer sep=0pt, scale=  1.96] at ( 48.72,296.48) {0.6};

\node[text=drawColor,rotate= 90.00,anchor=base,inner sep=0pt, outer sep=0pt, scale=  1.96] at ( 48.72,366.74) {0.8};

\node[text=drawColor,rotate= 90.00,anchor=base,inner sep=0pt, outer sep=0pt, scale=  1.96] at ( 48.72,437.01) {1.0};
\end{scope}
\end{tikzpicture}} \\ 
 (a) Mean $Acc_k$ vs. $k$ &
 (b) Mean $Err_k$ vs. $k$ &
 (c) Mean $\Delta_k$ vs. $k$ \\
\end{tabular}
\end{center}
\caption{Performance of each algorithm against $k$. (a) mean accuracy (b) mean Hausdorff distance (c) the relative error reduction.}
\label{fig:three}
\end{figure*}
\section{Related Work}

The problem of generating a computer program from some specification has been
studied since the beginnings of AI. Relevant work here falls into the two broad
camps of \emph{synthesis} where an explicit program is generated and
\emph{induction} where a latent representation may be used to generate input
output pairs \cite{devlin2017robustfill}.

In the inductive setting, there is a large literature of work relating to the
search for algorithms which correct a sketch. \cite{patidar2017correcting} uses
recurrent neural networks to model conditional sketch generation, an
image-to-image transformation problem. \cite{lake15science} uses probabilistic
program induction to perform one-shot modeling. In the programming synthesis
community, there is a large body of work synthesizing program from logical
specifications \cite{srinivasan2015synthesis}.

Synthesis settings can further be distinguished by whether a given specification
is \emph{partial} or \emph{total}. Abstractly, we may think of synthesis as
attempting to infer some $f \in \mathcal{F}$ where $\mathcal{F}$ is a family of
programs taking inputs from $\mathcal{X}$ and outputting $\mathcal{Y}$. A
partial specification comes in the form of a set of input-output pairs
$\{(x_1,y_1),\dots,(x_n,y_n)\}$ whose input items are a proper subset of
$\mathcal{X}$. This induces a problem of inference as the synthesis algorithm
must settle on some choice of output for unobserved input. By way of contrast, a
total specification fully specifies $f$ as a function. But even if we know the
specification, we may have difficulty finding a matching program. The total
specification synthesis problem is a problem of search rather than inference.

A further distinction can be made between settings where the specification is
\emph{noisy} or \emph{noiseless}. More recent work such as
\cite{devlin2017robustfill} and \cite{murray2016probabilistic} naturally handle
noisy specifications owing to their use of neural models.

\cite{gaunt2016terpret} and \cite{riedel2016programming} allow the user to
sketch a partial program in addition to a specification through inputs/output
pairs. Our method is distinct in that the user provides not a sketch but a
partial program which may contain errors. That is, our synthesis is not
constrained by the given partial program.

\section{Results}

We compute $k$-ahead completions for $k=1,\ldots,6$. To ensure practical run times,
we allot a state budget of $b = 50,000$ programs for each algorithm and $k$. Likewise, 
we enforce a static horizon of $C = 6$ so that $\textrm{cost}(\testp^{(i)}, \hat{\testp}^{(i)}) \le 6$
for all completions.

\textbf{Figure}~\ref{fig:three} (a) plots mean $Acc_k$ for each algorithm against $k$. For
small $k$ such as $k=1$, IDPS will always recover $\testp^{(i)}$ unless a better fit
is nearby, while sampling search manages a less reliable 63\% recovery
rate.
Under our state budget constraint, IDPS's performance decreases almost
monotonically because it exhausts its budget before exploring deep states. By
contrast, sampling search distributes its exploration equally across all depths,
and consequently scales well to large $k$. Viewed another way, the best IDPS can
do for small $k$ is recover the original trajectory. \textbf{Figure}~\ref{fig:three} (b)
and \textbf{Figure}~\ref{fig:three} (c), which plot mean $Err_k$ and mean $\Delta_k$
respectively against $k$, reflect that sampling search has greater opportunity
to explore deeper and more structured programs.

The dashed line of \textbf{Figure}~\ref{fig:three} (b) represents the mean
Hausdorff distances of the user's true completion $p^{(i)}$, that is, the mean of
$d_H(I(p^{(i)}),t^{(i)})$ across our corpus. As we regard the trajectory $t^{(i)}$ as the
true label, we can see that for smaller values of $k$ our algorithms improved
upon the user's completed program, $p^{(i)}$. For small lookaheads, this suggests
that users may benefit from viewing programs returned by our synthesis method.

Nonuniform sampling search dominates the sampling search regime, outperforming
uniform sampling in $Err_k$ and $\Delta_k$ for all $k$. By localizing block
targets in a manner statistically more consistent with observed human behavior,
the algorithm restricts its search to programs produced by high-level groupings
of commands, such as adding a turn block, then immediately connecting it to the
penultimate block, and then changing its angle parameter. This appears to
produce more reliable completion candidates.

It is worth noting that both uniform and nonuniform sampling search converge on better-fit trajectories when faced
with idiosyncratic programming techniques by the programmer. For example, one participant wrote a large loop body 
and only added the loop itself as his last step, while most other programmers added the loop first.
Under this setup, the block being connected inside the loop body is not local (in fact it is as distant as possible),
so the nonuniform algorithm is unlikely to sample an essential command. Nevertheless, the algorithm
produced a better Hausdorff fit through an altogether different program, an indication of robustness in our method.

\begin{figure}[t]
\begin{center}
\resizebox{0.25\textwidth}{!}{
\begin{tikzpicture}[x=1pt,y=1pt]
\definecolor{fillColor}{RGB}{255,255,255}
\path[use as bounding box,fill=fillColor,fill opacity=0.00] (0,0) rectangle (505.89,505.89);
\begin{scope}
\path[clip] (  0.00,  0.00) rectangle (505.89,505.89);
\definecolor{drawColor}{RGB}{0,0,0}
\definecolor{fillColor}{RGB}{0,0,139}

\path[draw=drawColor,line width= 0.4pt,line join=round,line cap=round,fill=fillColor] ( 83.76, 89.16) rectangle (193.16,437.01);
\definecolor{fillColor}{RGB}{255,0,0}

\path[draw=drawColor,line width= 0.4pt,line join=round,line cap=round,fill=fillColor] (215.04, 89.16) rectangle (324.45,310.61);
\definecolor{fillColor}{RGB}{255,255,0}

\path[draw=drawColor,line width= 0.4pt,line join=round,line cap=round,fill=fillColor] (346.33, 89.16) rectangle (455.73,267.68);
\end{scope}
\begin{scope}
\path[clip] (  0.00,  0.00) rectangle (505.89,505.89);
\definecolor{drawColor}{RGB}{0,0,0}

\node[text=drawColor,anchor=base,inner sep=0pt, outer sep=0pt, scale=  1.96] at (138.46, 55.44) {idps};

\node[text=drawColor,anchor=base,inner sep=0pt, outer sep=0pt, scale=  1.96] at (269.75, 55.44) {uniform};

\node[text=drawColor,anchor=base,inner sep=0pt, outer sep=0pt, scale=  1.96] at (401.03, 55.44) {nonuniform};
\end{scope}
\begin{scope}
\path[clip] (  0.00,  0.00) rectangle (505.89,505.89);
\definecolor{drawColor}{RGB}{0,0,0}

\node[text=drawColor,rotate= 90.00,anchor=base,inner sep=0pt, outer sep=0pt, scale=  2.24] at ( 15.12,261.34) {Mean running time (s)};
\end{scope}
\begin{scope}
\path[clip] (  0.00,  0.00) rectangle (505.89,505.89);
\definecolor{drawColor}{RGB}{0,0,0}

\path[draw=drawColor,line width= 0.4pt,line join=round,line cap=round] ( 68.88, 89.16) -- ( 68.88,397.93);

\path[draw=drawColor,line width= 0.4pt,line join=round,line cap=round] ( 68.88, 89.16) -- ( 60.48, 89.16);

\path[draw=drawColor,line width= 0.4pt,line join=round,line cap=round] ( 68.88,150.91) -- ( 60.48,150.91);

\path[draw=drawColor,line width= 0.4pt,line join=round,line cap=round] ( 68.88,212.67) -- ( 60.48,212.67);

\path[draw=drawColor,line width= 0.4pt,line join=round,line cap=round] ( 68.88,274.42) -- ( 60.48,274.42);

\path[draw=drawColor,line width= 0.4pt,line join=round,line cap=round] ( 68.88,336.17) -- ( 60.48,336.17);

\path[draw=drawColor,line width= 0.4pt,line join=round,line cap=round] ( 68.88,397.93) -- ( 60.48,397.93);

\node[text=drawColor,rotate= 90.00,anchor=base,inner sep=0pt, outer sep=0pt, scale=  1.96] at ( 48.72, 89.16) {0};

\node[text=drawColor,rotate= 90.00,anchor=base,inner sep=0pt, outer sep=0pt, scale=  1.96] at ( 48.72,150.91) {100};

\node[text=drawColor,rotate= 90.00,anchor=base,inner sep=0pt, outer sep=0pt, scale=  1.96] at ( 48.72,212.67) {200};

\node[text=drawColor,rotate= 90.00,anchor=base,inner sep=0pt, outer sep=0pt, scale=  1.96] at ( 48.72,274.42) {300};

\node[text=drawColor,rotate= 90.00,anchor=base,inner sep=0pt, outer sep=0pt, scale=  1.96] at ( 48.72,336.17) {400};

\node[text=drawColor,rotate= 90.00,anchor=base,inner sep=0pt, outer sep=0pt, scale=  1.96] at ( 48.72,397.93) {500};
\end{scope}
\end{tikzpicture}}
\end{center}
\vspace*{-5mm}
\caption{Mean running time (s) per algorithm}
\label{fig:runtime}
\end{figure}
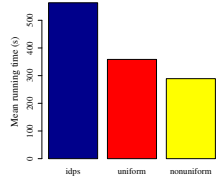

\textbf{Figure}~\ref{fig:runtime} shows that sampling search is consistently faster than IDPS,
the nonuniform (resp. uniform) variant requiring an average of 5.97 minutes (resp. 4.82 minutes) 
for convergence as opposed to 9.83 minutes.
That the former is nearly twice as fast as the latter is not surprising. For both the uniform
and nonuniform variant, sampling search is bottlenecked by the quadratic Hausdorff computation,
while IDPS, on top of Hausdorff, must expand many useless states.

\begin{figure}[htb]
\begin{center}
\begin{tabular}{cc}
\includegraphics[width=0.14\textwidth]{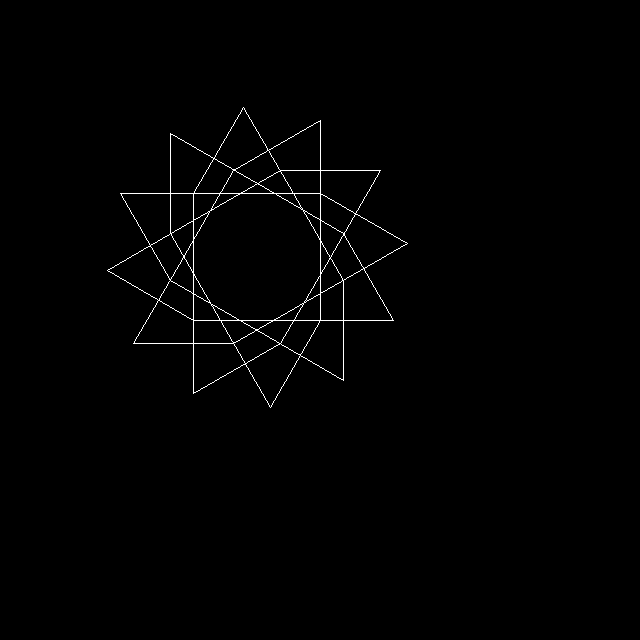} & \includegraphics[width=0.14\textwidth]{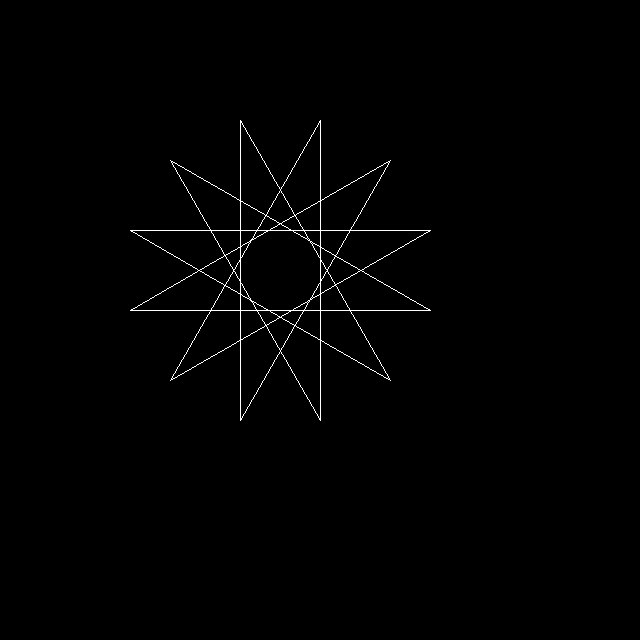} \\
  $k=3$, uniform & $k=4$, uniform \\
\includegraphics[width=0.14\textwidth]{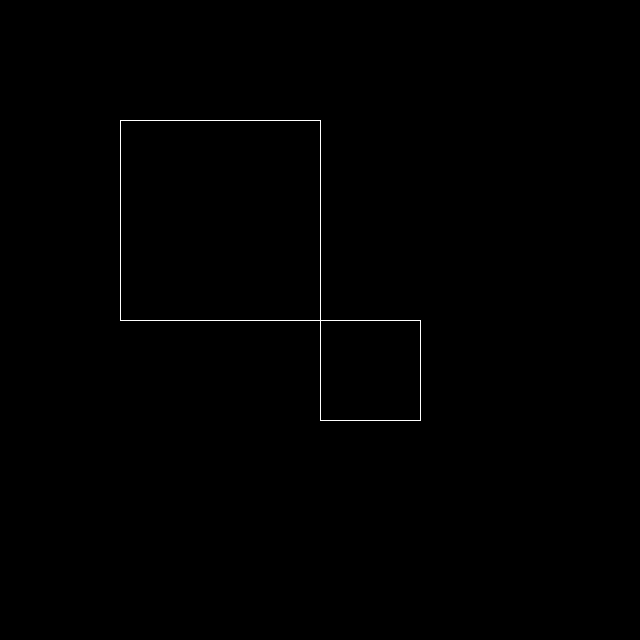} & \includegraphics[width=0.14\textwidth]{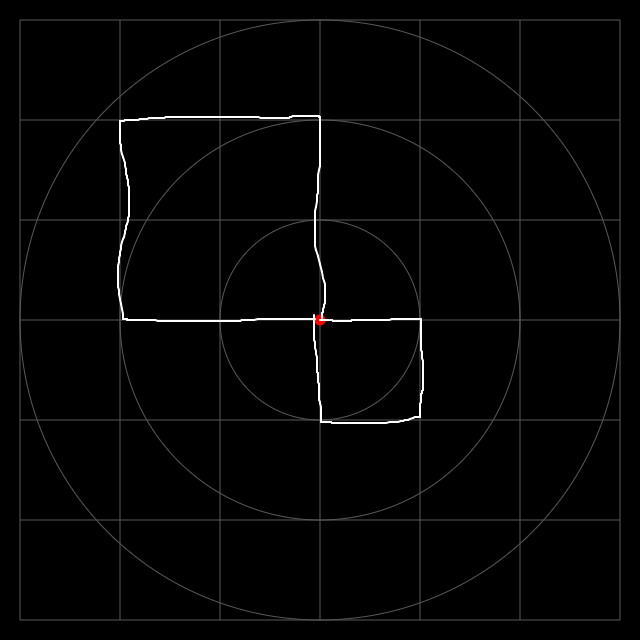} \\
  $k=3,4$, nonuniform & user trajectory \\
\end{tabular}
\end{center}
\caption{Trajectories drawn by participants and uniform, nonuniform solutions}
\label{fig:eh15}
\end{figure}

A qualitative look at some specific results can further insight. For example,
consider item 18 from our corpus as presented in \textbf{Figure}~\ref{fig:eh15}.
Here we show in the top row the solution obtained by the uniform algorithm for
lookaheads $k = 3$ and $k = 4$ and in the bottom row the user-drawn trajectory
and the nonuniform solution, which was the same in both cases of $k$. As we can
see the uniform algorithm returns a solution which ``cheats'' with respect to
our metric, constructing a trajectory which covers the region, without really
approximating it. The Hausdorff distance of the $k = 3$ and $k = 4$ solutions in
the top row is 89 and 90, respectively. By way of contrast, the nonuniform
algorithm found the user's solution, which had a distance to the specification
trajectory of 9. We conjecture that two factors contributed to the poor
performance of the uniform algorithm. First, intuitively, there are a small
number of ways to correctly complete the program, but a vast number of ways to
construct spirals of the form found by the uniform algorithm. Second, the user's
partial program, the starting point of search, was characterized by a large
number of \texttt{repeat} blocks, from which spiral solutions of the form found
were plentiful in the search space, as attachments which tended to nest
\texttt{repeat}s tend to draw such trajectories. The nonuniform solution,
modeling as it does the attention or focus of the programmer, overcame these
shortcomings.

\begin{figure}[htb]
\begin{center}
\begin{tabular}{ccc}
\includegraphics[width=0.14\textwidth]{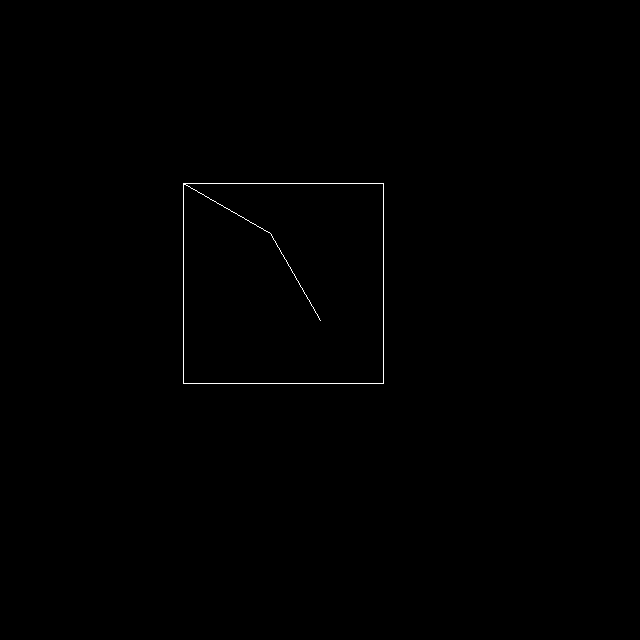} & \includegraphics[width=0.14\textwidth]{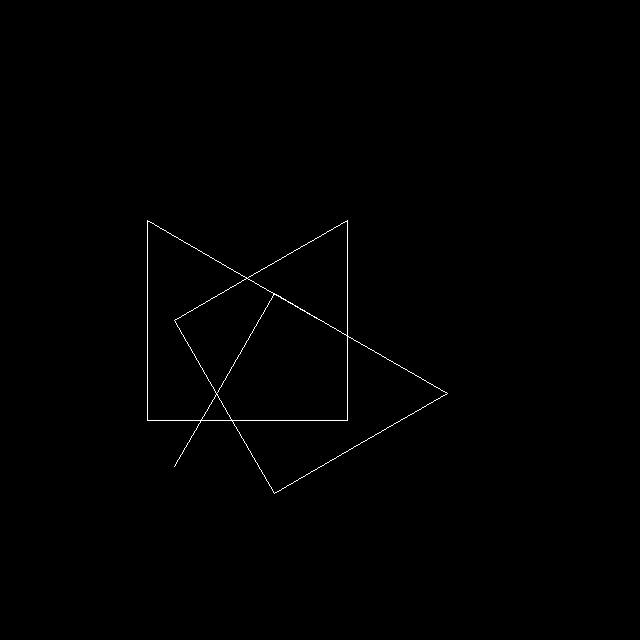} & \includegraphics[width=0.14\textwidth]{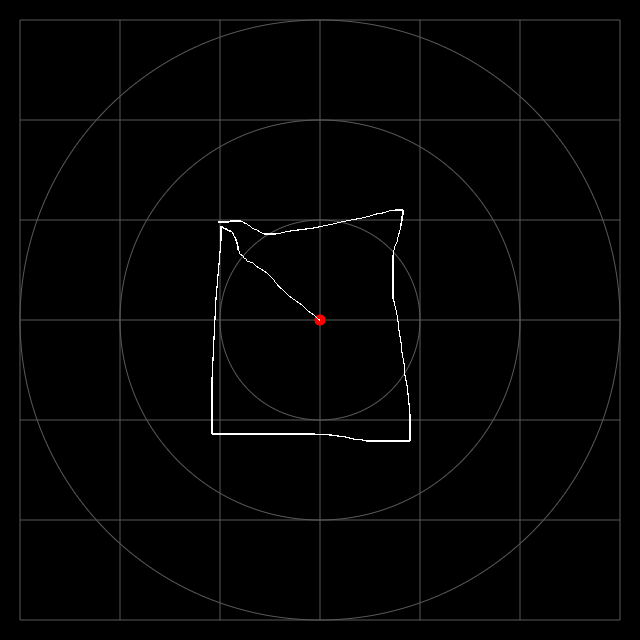} \\
  $k=6$, uniform & $k=6$, nonuniform & user trajectory
\end{tabular}
\end{center}
\caption{Trajectories drawn by participants and uniform, nonuniform solutions}
\label{fig:av3}
\end{figure}

Nevertheless, the \texttt{nonuniform} algorithm did not perform strictly better.
For example, item 3 in our corpus is presented in \textbf{Figure}~\ref{fig:av3}. Here,
the uniform algorithm returned a reasonable solution, while the nonuniform
seemed to lose its way. This example is notable for another reason in that it
demonstrates how the Hausdorff distance may not encode all features of the
user's intention. We see here that the program returned by the
\texttt{uniform} algorithm adds a bend to the line inside the square, while the
user's trajectory is suggests a straight curve. This raises the following
question: is the angle of the line within the square the more salient feature of
the user's intention or its straightness? The Hausdorff distances ``fudges'' by
adding a curve, while it could be argued the trajectory suggests that the
particular angle is less important than the fact that the line is straight.

As the quantitative results show, however, the \texttt{nonuniform} algorithm
fared significantly better. We wish to emphasize the conceptual significance of
this better performance. From a statistical point of view, we would argue
that our corpus is not drawn from the ``true distribution'' for our task, namely
some kind of distribution arising from expert performance. So we may wonder if
such a distribution would even be helpful in guiding a search a algorithm. This
can be framed as a tradeoff: if we hew too closely to the observed behavior of
users struggling to complete a task, we could overfit their idiosyncrasies and
blind spots, and our algorithm could inherit their limitations. If we ignore
their focus entirely and do not guide our search by some knowledge of how
programs are written, our search would be too uninformed to perform well. What
our succession of models suggests is that there is still enough statistical
signal even in the work of novice programmers to guide and improve search.

\section{Conclusion}

We have formulated program synthesis with visual specification in the frame of
classical AI search and have proposed two algorithms. Sampling methods produce
improved solutions and scale more readily to larger problem instances. A
sampling method informed by the attention and distribution of behaviors observed
from novice programmers leads to further improvement. We demonstrated that these
algorithms can outperform humans at their own intended tasks for smaller
lookahead values, suggesting a practical benefit for a synthesis method that can
complete a user's programs.

\section*{Acknowledgments}

We would like to thank Scott Alfeld for insightful comments and discussion. This
work is supported by NSF grant 1423237.

\bibliographystyle{plain}
\bibliography{arxiv}

\end{document}